\UseRawInputEncoding

\documentclass[10pt,journal,compsoc]{IEEEtran}
\usepackage{graphicx}
\usepackage{multirow}
\usepackage{subfigure}
\usepackage{float}

\ifCLASSOPTIONcompsoc
  \usepackage[nocompress]{cite}
\else
  \usepackage{cite}
\fi
\ifCLASSINFOpdf
\else
\fi
\begin{document}
%
\title{Neighborhood Spatial Aggregation MC Dropout for Efficient Uncertainty-aware\\Semantic Segmentation in Point Clouds}
%
%
%
%

\author{Chao~Qi
        and~Jianqin~Yin,~\IEEEmembership{Member,~IEEE}
\IEEEcompsocitemizethanks{\IEEEcompsocthanksitem Chao Qi is with the School of Artificial Intelligence, Beijing
University of Posts and Telecommunications, Beijing 100876, China, he is also with the Standard and Metrology Research Institute of China Academy of Railway Sciences Corporation Limited, Beijing 100081, China. \protect\\
E-mail: qichao199@163.com
\IEEEcompsocthanksitem Jianqin Yin (corresponding author) is with the School of Artificial Intelligence, Beijing
University of Posts and Telecommunications, Beijing 100876, China. \protect\\
E-mail: jqyin@bupt.edu.cn}
}

%
%

\markboth{Journal of \LaTeX\ Class Files,~Vol.~14, No.~8, August~2015}%
{Shell \MakeLowercase{\textit{et al.}}: Bare Demo of IEEEtran.cls for Computer Society Journals}
%



\IEEEtitleabstractindextext{%
\begin{abstract}
Uncertainty-aware semantic segmentation of the point clouds includes the predictive uncertainty estimation and the uncertainty-guided model optimization. One key challenge in the task is the efficiency of point-wise predictive distribution establishment. The widely-used MC dropout establishes the distribution by computing the standard deviation of samples using multiple stochastic forward propagations, which is time-consuming for tasks based on point clouds containing massive points. Hence, a framework embedded with NSA-MC dropout, a variant of MC dropout, is proposed to establish distributions in just one forward pass. Specifically, the NSA-MC dropout samples the model many times through a space-dependent way, outputting point-wise distribution by aggregating stochastic inference results of neighbors. Based on this, aleatoric and predictive uncertainties acquire from the predictive distribution. The aleatoric uncertainty is integrated into the loss function to penalize noisy points, avoiding the over-fitting of the model to some degree. Besides, the predictive uncertainty quantifies the confidence degree of predictions. Experimental results show that our framework obtains better segmentation results of real-world point clouds and efficiently quantifies the credibility of results. Our NSA-MC dropout is several times faster than MC dropout, and the inference time does not establish a coupling relation with the sampling times. The code will be available if the paper is accepted.
\end{abstract}

}

\maketitle

\IEEEdisplaynontitleabstractindextext

%
\IEEEpeerreviewmaketitle

\IEEEraisesectionheading{\section{Introduction}\label{sec:introduction}}

%
%
%
%
\IEEEPARstart{U}{ncertainty-aware} point cloud semantic segmentation (PCSS) focuses on two key missions: \textbf{uncertainty estimation} of segmentation results and \textbf{uncertainty-guided learning} for optimizing the segmentation model. Uncertainty estimation helps us know how much we can trust the predicted label of points, which is essential for decision-making applications, such as robotic grasping, path planning, and self-driving. Uncertainty-guided learning leverages the uncertainty to achieve better convergence of the model, aiming at improving segmentation performance.

\begin{figure}[ht]
    \centering
    \includegraphics[width=0.47\textwidth]{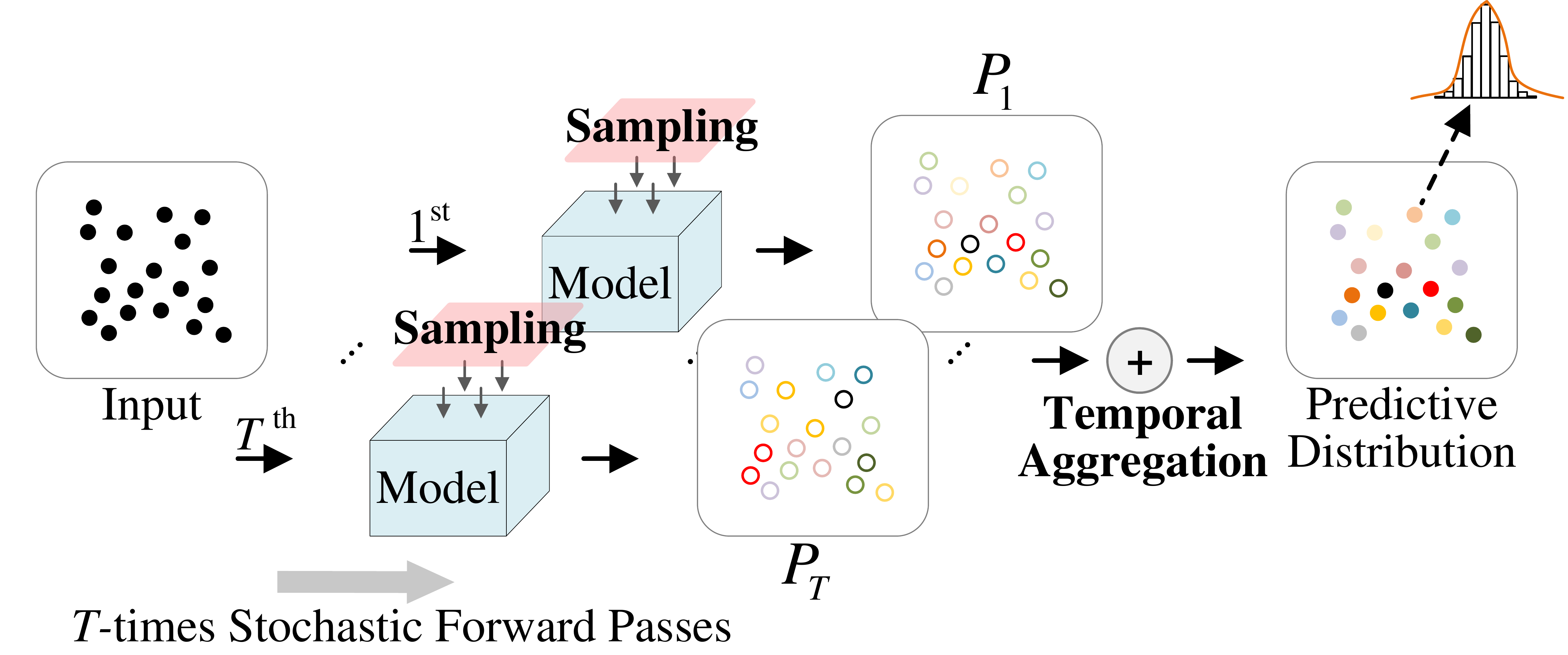}
    \caption{MC dropout samples the model by adding dropout layers in the backbone model to achieve the stochastic inference of point cloud. It repeats \emph{T} times to obtain \emph{T}-sets of probabilistic outputs $P_t$ for predictive distribution establishment, which results in \emph{T}-fold slower.}
    \label{fig1}
\end{figure}

There are lines of works considering general methods for uncertainty estimation \cite{Denker1991transforming,MacKay1992A,Neal1995,Gal2016,BlumHP15} and uncertainty-guided learning \cite{Kendall2017,feng2019leveraging,yi2019probabilistic,letham2019constrained,yu2019uncertainty,ebrahimi2020uncertainty,xia2020uncertainty,wang2020double,LiuLCHH20}. The key of the uncertainty-aware tasks is to model every target's predictive distribution to quantify different uncertainties. Nevertheless, these methods bring a large amount of computation, thus hindering their popularizations in uncertainty-aware PCSS. For uncertainty estimation, \cite{Denker1991transforming,MacKay1992A,Neal1995} propose a Bayesian model to establish every target's predictive distribution by learning (Gaussian) posterior probabilities over the weights. \cite{Gal2016,BlumHP15} have proved that dropout in NNs can be interpreted as an approximation to the Bayesian model. This widely used method, called Monte Carlo (MC) dropout, establishes each target's predictive distribution through aggregating repeated stochastic variational inferences, as shown in Fig. \ref{fig1}. This time-consuming process needs to perform stochastic forward passes through the model many times, which is harmful to tasks with high-efficiency requirements.

For uncertainty-guided learning, different decompositions of uncertainty play different roles in model training. The aleatoric component reflects the inherent observations' noise, and some works rewrite the loss function with this component to guide model optimization \cite{Kendall2017,feng2019leveraging,yi2019probabilistic,letham2019constrained}. Targets with high aleatoric are assigned low weights in the loss function to penalize the inherent noise for better model performance. Some works leverage the epistemic component to enhance the data cognition of the model in semi-supervised and continual learning tasks \cite{yu2019uncertainty,ebrahimi2020uncertainty,xia2020uncertainty,wang2020double,LiuLCHH20}. No matter which uncertainty component is used for the model optimization, it acquires from the predictive distribution. The uncertainty-guided learning methods overly depend on the MC dropout or additional variance parameter introductions, which are difficult to be generalized.

\begin{figure}[ht]
    \centering
    \includegraphics[width=0.47\textwidth]{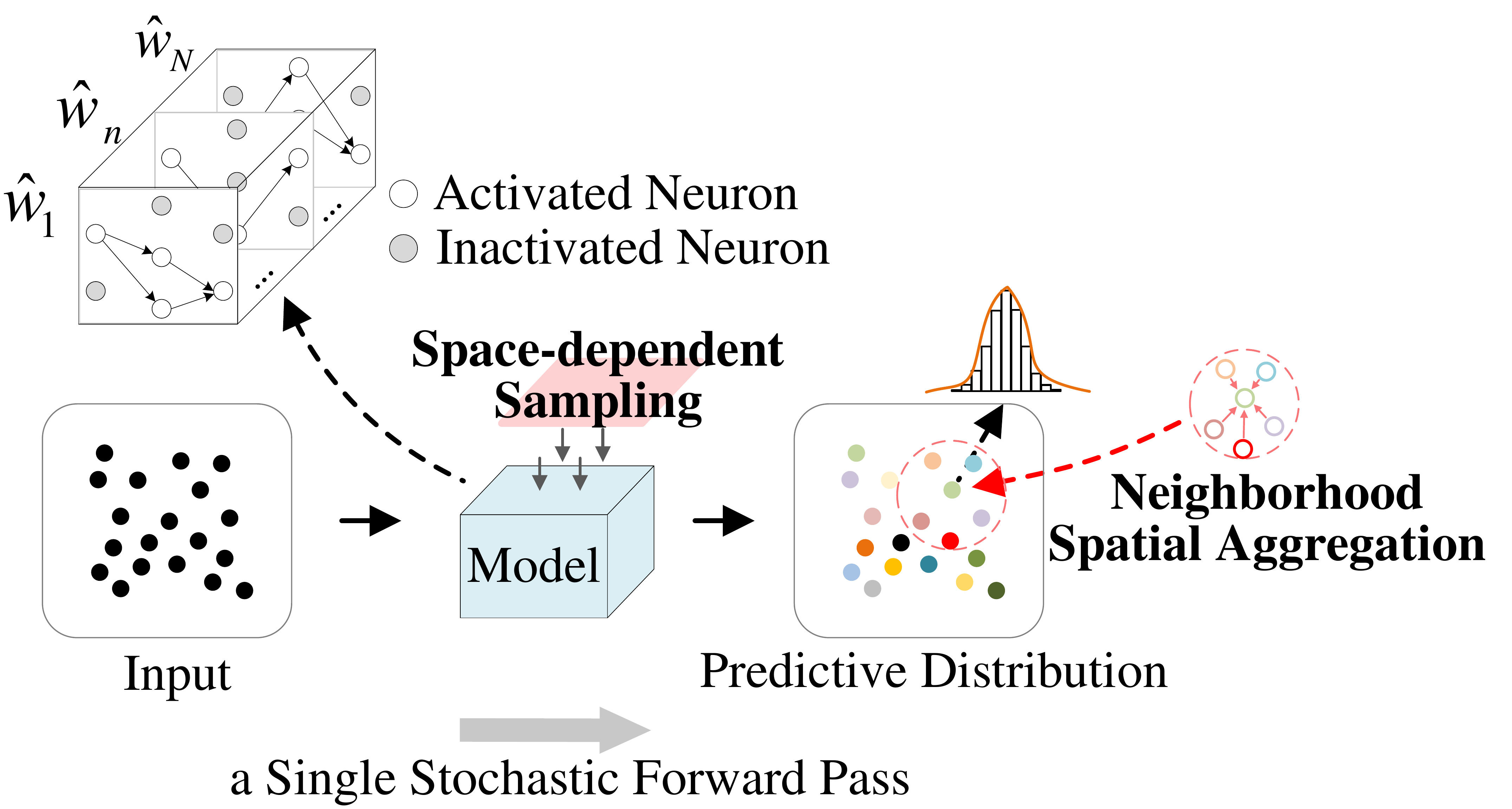}
    \caption{NSA-MC dropout samples the weight-sharing PCSS model in a single stochastic forward pass, resulting in \emph{N}-sets of random parameters ${{\hat{w}}_{{}}}$ corresponding to \emph{N} different spatial points (thus called space-dependent sampling). Point-wise predictive distribution is established through the aggregation of probabilistic outputs of neighborhood points.}
    \label{fig2}
\end{figure}

In summary, existing methods of uncertainty estimation and uncertainty-guided learning suffer from a time-consuming distribution establishment. For a point cloud generally containing millions or even tens of millions of points, these methods are inefficient for modeling point-wise predictive distribution. MC dropout is an easy-using method except for the repeated forward passes. It motivates us to explore a variant of MC dropout based on a single forward pass and can be used in efficient uncertainty-aware PCSS.

Unlike the 2D grid data, like images, the point cloud is a set of unordered points describing an object's geometric structure. Mainstream PCSS models independently work on each point and share weights across points to maintain the permutation invariance of the point cloud \cite{guo2020deep,charles2017pointnet,qi2017pointnet,hu2020randla,lei2020spherical,guo2021pct}. Thus, the stochastic inference of each point can be achieved by injecting point-wise randomness into the shared weights of the model. Besides, given the feature similarity of \emph{T}-nearest neighbors due to the local region's geometric continuity, the one-time stochastic inference of a point with neighbors can approximate the point's repeated stochastic inferences. Thus, each point's predictive distribution can be efficiently established by aggregating the probabilistic outputs of its neighborhood and well used in the uncertainty-aware PCSS.

Specifically, as shown in Fig. \ref{fig2}, we design a space-dependent sampling module to randomly drop units in the weight-shared backbone model, generating sets of randomly masked weights for the stochastic inference of different spatial points. Furthermore, a novel neighborhood spatial aggregation (NSA) module is proposed to establish each point's predictive distribution by aggregating probabilistic outputs of its neighborhood points. This novel predictive distribution establishment method based on a single forward propagation is called NSA-MC dropout.

Based on the efficient predictive distribution establishment, different types of uncertainties of each point are obtained using different acquisition functions. In the uncertainty estimation, the predictive uncertainty quantifies the predictive results' confidence degree without repeated inferences. In uncertainty-guided learning, considering the relationship between the aleatoric uncertainty and the dataset's inherent noises, we use the uncertainty as the weight to be integrated into the cross-entropy loss. This designed loss function prevents the model from over-fitting caused by the noise data, resulting in better model convergence and improved segmentation performance.

This work is a marked improved version of our previous studies presented in ICRA 2021 \cite{Chao2021}. Below, we summarize the major contributions extended beyond the preliminary work.

\begin{itemize}
  \item \textbf{An efficient uncertainty-aware PCSS framework} leverages uncertainty to improve the segmentation performance and quantify the confidence degree of predictive results. To our best knowledge, this is the first framework achieving uncertainty-aware point cloud semantic segmentation without increasing model parameters or repeated inferences.
  \item \textbf{Expanding use of the NSA-MC dropout}. Unlike \cite{Chao2021} only focuses on uncertainty estimation, we extend the novel distribution establishment method to uncertainty-guided learning. It quantifies the aleatoric uncertainty efficiently, penalizing inherent noises and promoting better model convergence.
  \item \textbf{Systematic evaluation on kinds of datasets}. Compare to \cite{Chao2021}, we conduct a more comprehensive experiment on real-world (including indoor and outdoor scenarios) and synthetic datasets to demonstrate the proposed framework. The presented results ascertain the high computational efficiency of our method with a segmentation performance improvement on benchmarks.
\end{itemize}

\section{Related work}

Even though there is little research on the uncertainty-aware PCSS, the general uncertainty-aware methods have been widely concerned in NN-based applications. We first review the related works of predictive distribution establishment. After that, we summarize the works related to uncertainty estimation and uncertainty-guided learning.

\subsection{Predictive Distribution Establishment}

In an uncertainty-aware neural network, the output is represented as a predictive distribution among all possible values, rather than having a single fixed value \cite{Charles2015}.

It isn't easy to obtain the actual output distribution of each target. \cite{Hinton1993,Barber1998,Graves2011,ebrahimi2020uncertainty} aim to approximate the distribution as a simple variational distribution, such as Gaussian. However, these methods increase model parameters, resulting in high demand for computing resources in many tasks, such as RGB-D saliency detection \cite{zhang2021uncertainty} and event prediction \cite{soleimani2018scalable}. The prediction variance generated by multiple stochastic inferences can be used to establish the predictive distribution \cite{Gal2016,BlumHP15}. This method, based on random model sampling, is called the MC dropout. It has been widely concerned due to its ease of use. However, MC dropout suffers from a time-consuming model sampling process relying on multiple forward passes despite its remarkable achievements. Thus, we explore a method to efficiently establish the point-wise predictive distribution without introducing additional parameters and repeated inferences.

\subsection{Uncertainty Estimation}

Different types of uncertainties capture from the predictive distribution using different acquisition functions. However, uncertainty estimation is limited by the inefficient distribution establishment. To address this problem, \cite{Postels2019ICCV} designed a sampling-free approach for uncertainty estimation, and injected noises are used to model the prediction variance for distribution establishment at test time. The method largely improves the uncertainty estimation efficiency, but the estimation performance depends on the designed noise layers. \cite{Huang2018ECCV} designed a region-based temporal aggregation (RTA) method for uncertainty estimation in video segmentation. It approximates the traditional temporal aggregation method in MC dropout by using the similarity of consecutive frames. Even though this proposed method cannot directly apply to PCSS, it inspires us to explore an approximate method of probabilistic result aggregation by using point cloud characteristics.

\subsection{Uncertainty-guided Learning}

Recently, some studies have explored utilizing uncertainty to guide model learning for prediction performance improvement. The uncertainties can be classified into two main categories: aleatoric and epistemic \cite{Kendall2017}. Aleatoric uncertainty captures the inherent noises in the dataset, while epistemic uncertainty quantifies the reliable degree of the model parameters. The epistemic uncertainty reduces by observing more data, but the aleatoric one cannot.

Noises can lead to model over-fitting and dramatically degrades the predictive performance \cite{yi2019probabilistic}. Based on the relationship between aleatoric uncertainty and inherent noise, \cite{Kendall2017} proposed integrating aleatoric uncertainty into the loss function for better noise robustness. Targets with high aleatoric will be weighted low in the loss function, resulting in worse fitting in noisy regions and better fitting in clean regions \cite{gurevich2019pairing}. This method improves model performance and has been used in lots of applications  \cite{Kendall2017,feng2019leveraging,yi2019probabilistic,letham2019constrained}. However, an obvious disadvantage of this method is that it needs to introduce parameters to model the predictive distribution used for aleatoric uncertainty estimation, thus causing higher demand for computing resources. To address this problem, we explore to model the aleatoric uncertainty without introducing additional parameters and design an easy-using uncertainty-guided loss function for PCSS.

In addition, epistemic uncertainty guiding the model optimization in semi-supervised learning and other tasks has been concerned. This kind of method also suffers from an inefficient predictive distribution establishment, and our findings can provide a reference to it.

\section{Approach}

\begin{figure*}[htb]
    \centering
    \includegraphics[width=0.7\textwidth]{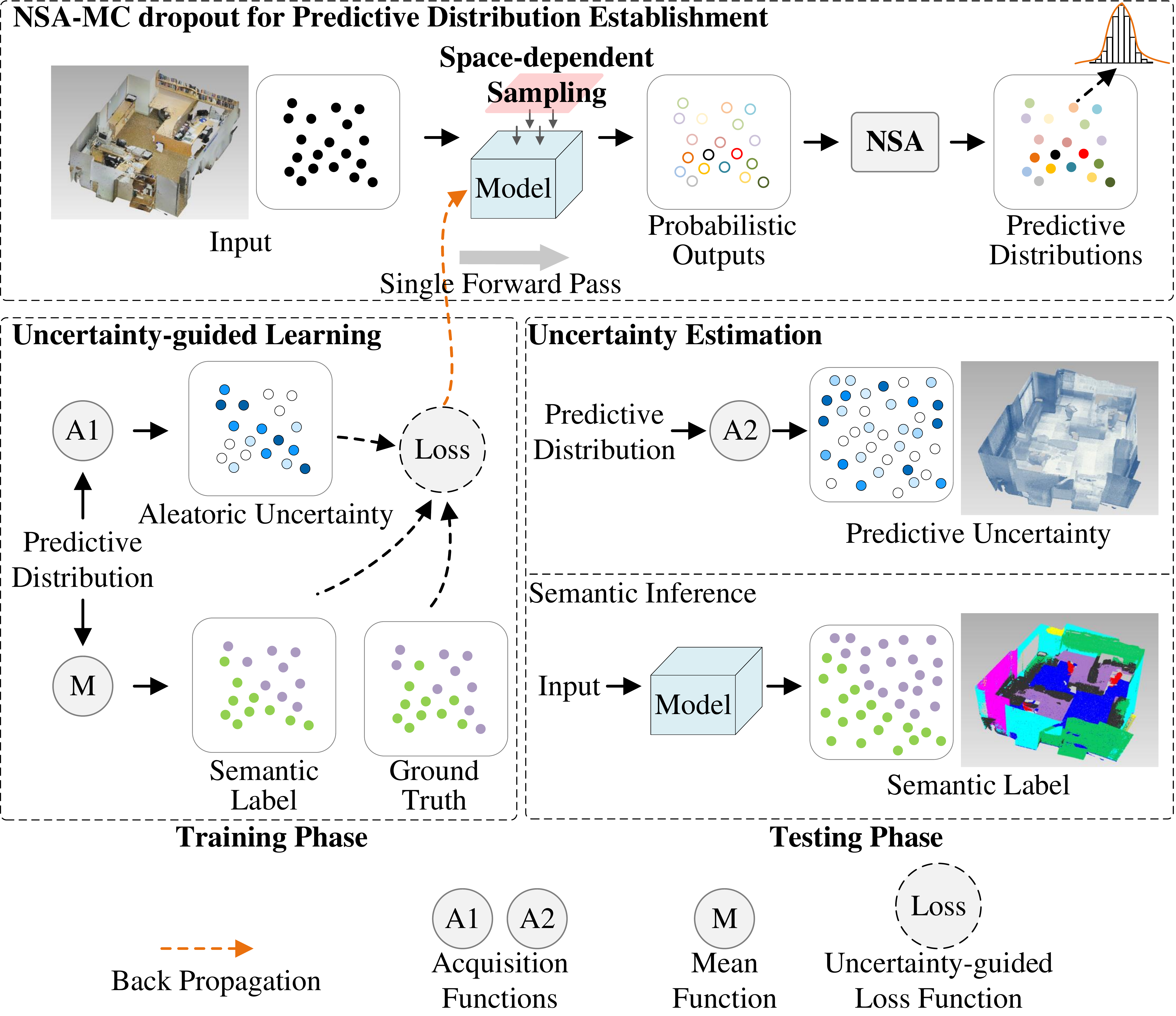}
    \caption{The overall uncertainty-aware PCSS framework. In predictive distribution establishment, the original point cloud is considered the input, and the predictive distribution of every point is considered the output. The space-dependent sampling module samples the backbone model many times in a single forward pass. The NSA module aggregates the local neighborhood probabilistic outputs for predictive distribution establishment. Taking points with predictive distribution as input, different types of uncertainty are obtained using different acquisition functions. In the training phase, points with high aleatoric uncertainty are assigned low weights in the cross-entropy loss for better model learning. In the testing phase, our framework outputs the improved segmentation results as well as the predictive uncertainty.}
    \label{fig3}
\end{figure*}

We propose a framework to achieve the efficient uncertainty-aware PCSS, as shown in Fig. \ref{fig3}. NSA-MC dropout (mainly including a space-dependent sampling module and an NSA module) establishes the predictive distribution in a single stochastic forward pass. The loss function, integrating the aleatoric uncertainty acquired from the predictive distribution, guides the backbone model for better convergence. Also, this framework outputs the predictive uncertainty to quantify the confidence degree of results.

\subsection{Predictive Distribution Establishment}

This section describes how the NSA module works with the space-dependent model sampling module to establish the predictive distribution in a single forward propagation.

\subsubsection{Neighborhood Spatial Aggregation}
Traditional MC dropout collects each point's \emph{T} random probabilistic outputs $\{ {p_t}(x)\} _{t = 1}^T$ as the empirical samples to denote the predictive distribution, as shown in Fig. \ref{fig1}. Each point's probabilistic output ${p_t}(x) \in {P_t}$ is obtained through a stochastic inference. ${{p}_{t}(x)}=p({{\hat{y}}_{t}}|x,{{\hat{w}}_{t}})$ denotes the \emph{t}-th probabilistic output while ${{\hat{y}}_{t}}$ is the predictive label and ${{\hat{w}}_{t}}$ represents the randomly masked weights generated by model sampling.

We design a spatial aggregation method to eliminate the dependence of predictive distribution establishment on \emph{T}-times forward propagations. For a set of \emph{T}-NN points $N(x)$, which contains $x$ and its \emph{T}-1 nearest neighbors (NN), the probabilistic output of included points is similar due to the consistency of geometric features. It can be denoted as:

\begin{equation}
p(y|{{x}^{i}},w)\approx p(y|{{x}^{j}},w)\ s.t.\ {{x}^{i}},{{x}^{j}}\in N(x)\label{1}
\end{equation}
Thus, \emph{T}-NN samples ${\{ {p_{t = 1}}{\rm{(}}{x^i}{\rm{)}}\} _{{x^i} \in N(x)}}$ can approximate the MC dropout samples $\{{{p}_{t}}(x)\}_{t=1}^{T}$ of predictive distribution, and the mean value of distribution can be expressed as:
\begin{equation}
\frac{1}{T}\sum\limits_{{x^i} \in N(x)} {p(\hat y|{x^i},{{\hat w}_i})}  \approx \frac{1}{T}\sum\limits_t {p({{\hat y}_t}|x,{{\hat w}_t})}\label{2}
\end{equation}
This process, shown in Fig. \ref{fig2}, is called neighborhood spatial aggregation. If $\{{{\hat{w}}_{1}},...,{{\hat{w}}_{T}}\}$ used for the stochastic inference of \emph{T} nearest points can be created in a single stochastic forward pass, the predictive distribution can also be established.

\subsubsection{Space-dependent Model Sampling}

Mainstream PCSS methods process each point identically and independently to maintain the permutation invariance of input. These backbone models share weights $w$ across \emph{T} nearest points. $\{{{\hat{w}}_{1}},...,{{\hat{w}}_{T}}\}$ can be generated by placing a Bernoulli matrix $\{{{z}_{1}},...,{{z}_{T}}\}$ over the matrix with shared weights $\{w,...,w\}$, where ${{z}_{t}}\sim Bernoulli(p)$ denoting the dropout process. Cooperating with the NSA module, the predictive distribution of each point denoted by empirical samples ${\{ {p_{t = 1}}{\rm{(}}{x^i}{\rm{)}}\} _{{x^i} \in N(x)}}$ is established in a single forward pass.

\begin{figure*}[htb]
    \centering
    \includegraphics[width=0.57\textwidth]{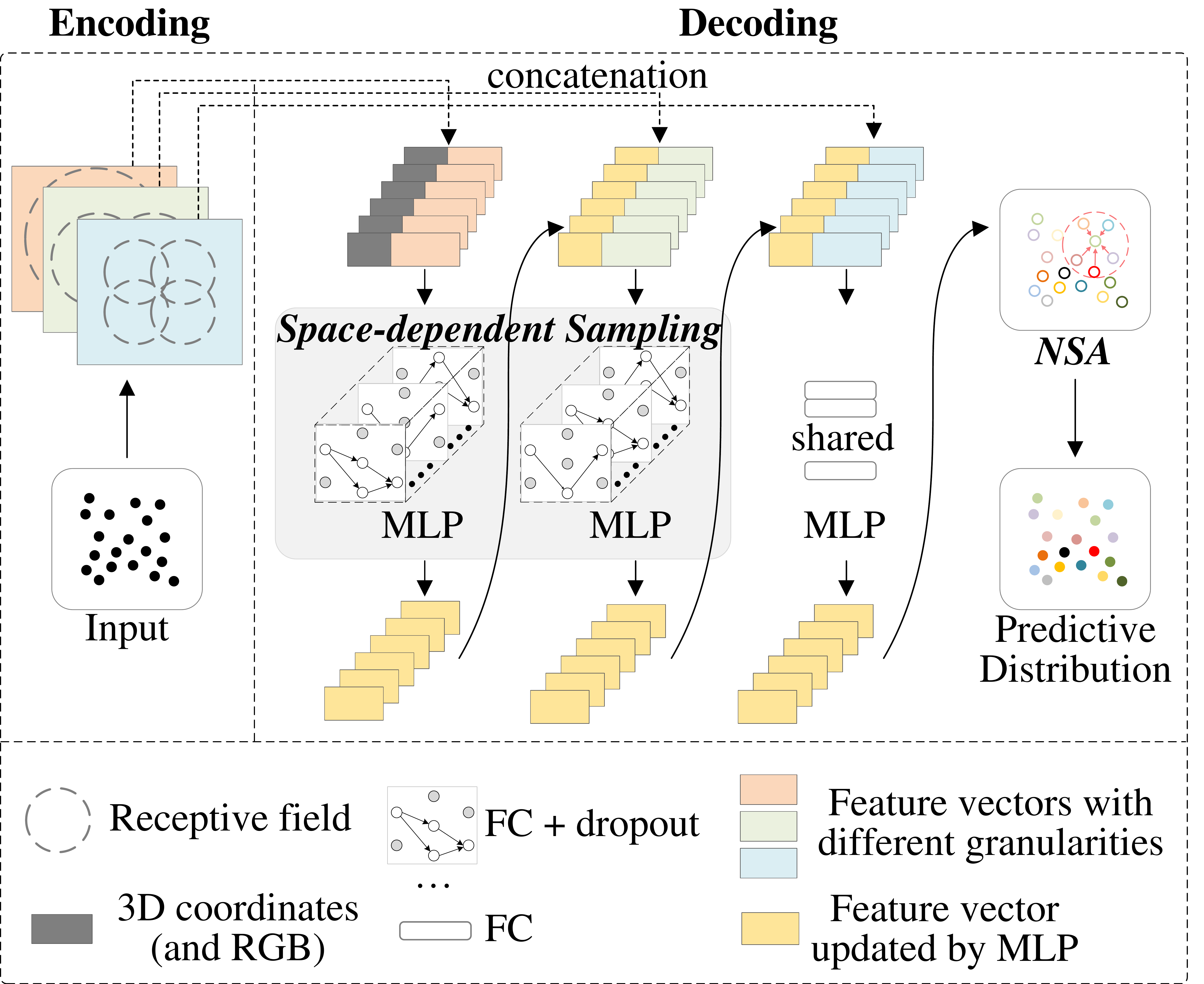}
    \caption{Uncertainty-aware PCSS with PointNet(++) as backbone. In the encoding stage, the network with multi-scale receptive fields generates feature vectors with different granularities. In the decoding stage, features are concatenated to points for further classification in a hierarchical fashion, and dropouts are inserted after MLPs to achieve model sampling. NSA outputs predictive distribution for the estimation of different types of uncertainties.}
    \label{fig2.5}
\end{figure*}

Considering the great success of PointNet \cite{charles2017pointnet} and PointNet++ \cite{qi2017pointnet} in point cloud segmentation, we choose them as backbone models. PointNet(++) is an encoder-decoder network, as shown in Fig. \ref{fig2.5}. The encoding part learns the global feature or local features. The decoding part concatenates the features to each point for further point-wise predictive distribution establishment.

A fully Bayesian network should be trained and tested with dropout after each trainable layer. \cite{Kendall2015} found that this was too strong a regulariser, leading to a slow convergence speed for model training. Besides, \cite{Kendall2015} explored many variants that only insert dropouts after partial layers for model sampling. These variants generate similar-looking uncertainty outputs as the fully Bayesian network.

The following issues should be considered to achieve effective model sampling. On the one hand, Multi-Layer Perceptron (MLP) used for decoding contains a few weight-sharing fully connected (FC) layers. Through randomly dropping units in MLP out, FCs used for the inference of different points present randomly masked weights, thus achieving the space-dependent model sampling in a single stochastic forward. On the other hand, encoding information shares among points in the same hierarchical group. Adding dropout layers in the encoding stage contributes nothing to the point-wise stochastic inference in a single forward pass. Thus, dropout layers are only inserted after MLPs in the decoding stage.

\subsection{Uncertainty Estimation and Decomposition}

\begin{table*}[ht]
\caption{Acquisition functions for uncertainty estimation and decomposition.}
\label{table1}
\begin{center}
\begin{tabular}{c|c}
\hline
Predictive entropy (PE)\\ \cite{Shannon2001,Gal2016phd,depeweg2018decomposition}& $\underbrace { - \sum\limits_c {(\frac{1}{T}\sum\limits_t {{p_{t,c}}(x)} )} \log (\frac{1}{T}\sum\limits_t {{p_{t,c}}(x)} )}_{ = :{\rm{ }}H \propto U} = \underbrace {{H_e}}_{ \propto {\rm{ }}{U_e}}\underbrace { - \frac{1}{T}\sum\limits_{c,t} {{p_{t,c}}(x)} \log {p_{t,c}}(x)}_{ = :{\rm{ }}{H_a} \propto {U_a}}$ \\ \hline
Mean Standard deviation (STD)\\ \cite{kampffmeyer2016semantic,kwon2020uncertainty}& $\underbrace {Var}_{ \propto {\rm{ }}U} = \underbrace {\frac{1}{T}\sum\limits_t {{{({p_{t,c}}(x) - \frac{1}{T}\sum\limits_t {{p_{t,c}}(x)} )}^{ \otimes 2}}} }_{ = :{\rm{ }}Va{r_e} \propto {U_e}} + \underbrace {\frac{1}{T}\sum\limits_t {[diag(\frac{1}{T}\sum\limits_t {{p_{t,c}}(x)} ) - {p_{t,c}}{{(x)}^{ \otimes 2}}]} }_{ = :{\rm{ }}Va{r_a} \propto {U_a}}$ \\ \hline
\end{tabular}
\end{center}
\end{table*}

Our efficient predictive distribution establishment method supports the estimation and decomposition of uncertainties using different acquisition functions, such as the \emph{predictive entropy} (PE) and the mean \emph{standard deviation} (STD). The PE-based function $H( \cdot )$ quantifies the uncertainty by capturing the amount of information contained in the predictive distribution \cite{Gal2016phd}, and the STD-based function $Var( \cdot )$ achieves the quantification by modeling the deviation of the distribution \cite{KampffmeyerSJ16,Kendall2015}.

With STD-based function as an example \cite{kwon2020uncertainty}, the aleatoric uncertainty ${U_a} \propto {Var_a}$ captures the vagueness inherent in the input dataset, and the epistemic uncertainty ${U_e} \propto {Var_e}$ refers to the model confidence. The combination of these two uncertainties presents the predictive uncertainty $U$ -- the model's confidence in the prediction results taking into account noise. ${Var_a}$ and ${Var_e}$ respectively represent the aleatoric and epistemic components of STD.

Uncertainties denoted by empirical samples are illustrated in Table \ref{table1}, where $p({\hat y_t}=c|x,{\hat w_t})$ is denoted as ${p_{t,c}}(x)$ for convenience. By using our NSA-MC dropout, $\frac{1}{T}\sum\limits_{{x^i} \in N(x)} {f({p_{t = 1,c}}({x^i}))}$ can approximate $\frac{1}{T}\sum\limits_t {f({p_{t,c}}(x))}$, where $f( \cdot )$ indicates the subfunction used in the acquisition function such as $\log ( \cdot )$. It makes the uncertainty estimation and decomposition achieved in a single forward pass.

\subsection{Uncertainty-guided Learning}

\begin{figure}[ht]
    \centering
    \includegraphics[width=0.47\textwidth]{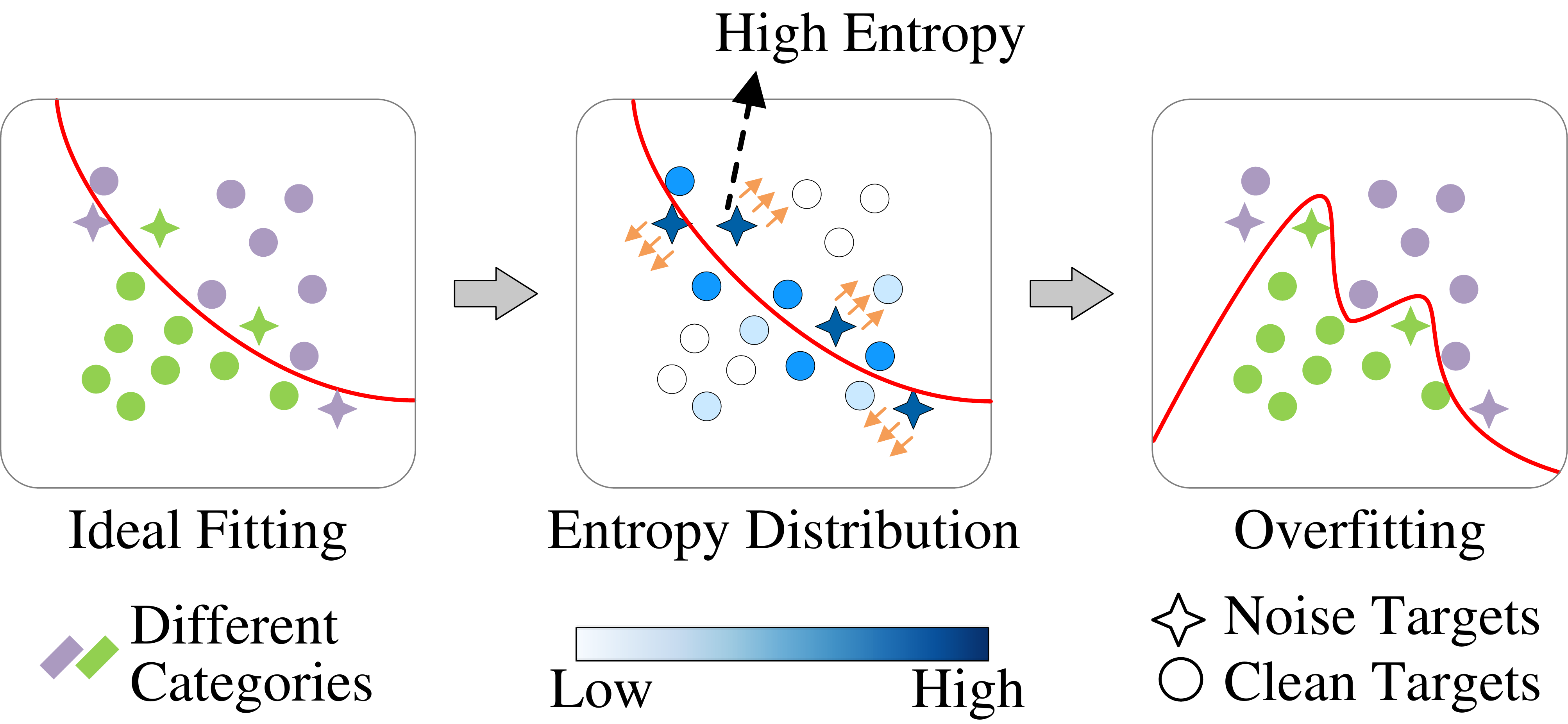}
    \caption{During model learning, noisy targets contribute a lot to the cross-entropy loss. It changes the trend of the fitting curve and easily leads the model to over-fitting.}
    \label{fig4}
\end{figure}

Generally, noise targets contribute a higher value to the loss during model learning, as shown in Fig. \ref{fig4}. It makes the model over-fitting and causes performance degradation. To address this problem, the predictive output of a target as a distribution is modeled among all possible values, which lends the flexibility to handle noise-contaminated and noise-free labels in a unified way \cite{bentaieb2017uncertainty}.

In PCSS, noisy points generally have a large predictive variance. Noises are assigned low weights in the uncertainty-guided loss function to achieve stronger noise robustness . The function is defined in (\ref{3}), where \emph{N} and \emph{M} represent the number of points and classes, respectively, and ${{\sigma }^{i}}$ denotes a measure of predictive variance given by aleatoric uncertainty $U_{a}^{i}$. ${{\beta }_{i}}_{,c}=1$ if \emph{c} equals the true label, else it is 0. The first term of uncertainty-guided loss function tends to fit worse on noisy points and uses this freedom to fit better on certain and clean points. The uncertainty regularization term $\log ({{\sigma }^{i}})$ prevents the network from predicting infinite uncertainty (and therefore zero loss) for all points.
\begin{equation}
L=\frac{1}{N}\sum\limits_{i=1}^{N}{[\frac{1}{{{\sigma }^{i}}}\sum\limits_{c=1}^{M}{-{{\beta }_{i,c}}}\log p(y=c|{{x}^{i}},w)+\log ({{\sigma }^{i}})]}\label{3}
\end{equation}
The uncertainty-guided learning penalizes the inherent noise in observations. However, it introduces an additional learnable variance ${{\sigma }^{i}}$. Our proposed method, as discussed above, addresses this problem and quantifies aleatoric uncertainty $U_{a}^{i}\propto {{\sigma }^{i}}$ in a single stochastic forward without adding parameters. We rewrite the loss function with aleatoric uncertainty as followed.
\begin{equation}
L=\frac{1}{N}\sum\limits_{i=1}^{N}{\frac{1}{1+\alpha U_{a}^{i}}\sum\limits_{c=1}^{M}{-{{\beta }_{i,c}}}\log p(y=c|{{x}^{i}},w)}\label{4}
\end{equation}
The uncertainty regularization term is abandoned because $U_{a}^{i}$ is not a learnable parameter. Besides, considering the ambiguous boundary quantified by aleatoric uncertainty between the noisy points and clean points, we map $1/U_a^i$ into a narrow range $1/(1 + \alpha U_a^i)$ to prevent underfitting of clean points, where $\alpha$ is the weighting value.

\section{Experiment}
\subsection{Dataset}

We evaluate the performance of our method for semantic segmentation and uncertainty estimation using indoor and outdoor, real-world and synthetic datasets, including S3DIS \cite{Iro2016}, NPM3D \cite{roynard2018paris}, and ShapeNet \cite{yi2016a}.

\textbf{S3DIS} is a set of large-scale point clouds collected in real-world indoor scenes. It comprises colored point clouds collected for 271 rooms belonging to 6 large-scale indoor areas in 3 buildings using 3D scanners. 13 class labels are included in this dataset. Each point cloud in the dataset corresponds to a single room. Point clouds in Area-5 are used for testing, while others for training. It is difficult for the backbone PointNet(++) to directly process the point cloud of a single room. Thus, every point cloud is split into 1m by 1m blocks for further processing. Backbone models are trained on 4096 sample points in each block.

\textbf{NPM3D} is a large-scale urban outdoor point cloud dataset acquired by a mobile laser scanning system in two cities in France: Lille and Paris. This dataset covers about 2 km of streets, containing more than 140 million points with 10 coarse classes and 50 fine classes. We consider XYZ coordinates and the reflectance value as input features and ignore other provided information, such as GPS time and the position of the LiDAR. Two scenes, Lille1 and Lille2, are used for training while the scene Paris for testing. Similar to S3DIS, NPM3D is processed in a sliding window fashion with blocks of 8m by 8m. Cropped scenes containing 8192 sample points are used for training.

\textbf{ShapeNet} is a set of point clouds of synthetic objects created by collecting CAD models from online open-sourced 3D repositories. It contains 16,881 synthetic models from 16 categories. These models in each category have two to five annotated parts, for a total of 50 parts. This dataset provides XYZ coordinates, RGB values, and the normal vector as input features and has 14,007/2,874 training/testing split defined.

\subsection{Implementation Details}

\textbf{Uncertainty-guided model training}. The backbone PointNet(++) is end-to-end trained using back-propagation and Adam optimizer with an initial learning rate of 0.001. The training batch size is set to 16 in all the backbones. All the experiments are conducted by PyTorch framework on a single GeForce GTX 1080 Ti and an Intel(R) Core(TM) i7-6850K 3.60GHz 6 cores CPU. Dropouts are added after every decoder layer with learnable parameters in backbone models while training on ShapeNet and S3DIS. However, it leads to a too slow converge speed on NPM3D, and we reduce the dropout layers to address this problem. We use the KD-Tree algorithm to search the ten nearest neighbors in parallel for neighborhood spatial aggregation. Aleatoric uncertainty with a weight value of 0.5, discussed below, is integrated into the loss function.

\textbf{Predictive uncertainty estimation}. Dropouts keep open at testing time to achieve model sampling for uncertainty estimation. Acquisition functions illustrated in Table \ref{table1} quantify different types of uncertainties.

\begin{table*}[]
\centering
\caption{Quantitative results on different datasets.}
\label{table2}
\begin{tabular}{c|c|ccc|ccc|ccc}
\hline
\multirow{3}{*}{Backbone}   & \multirow{3}{*}{\begin{tabular}[c]{@{}c@{}}Uncertainty-\\ guided\\ learning\end{tabular}} & \multicolumn{3}{c|}{\multirow{2}{*}{S3DIS}}   & \multicolumn{3}{c|}{\multirow{2}{*}{NPM3D}}   & \multicolumn{3}{c}{\multirow{2}{*}{ShapeNet}} \\
                            &                                                                                           & \multicolumn{3}{c|}{}                         & \multicolumn{3}{c|}{}                         & \multicolumn{3}{c}{}                          \\ \cline{3-11}
                            &                                                                                           & mIoU (\%)     & mAcc (\%)     & oAcc (\%)     & mIoU (\%)     & mAcc (\%)     & oAcc (\%)     & mIoU (\%)     & mAcc (\%)     & oAcc (\%)     \\ \hline
\multirow{2}{*}{PointNet}   & w/o                                                                                       & 45.6          & 55.0          & 80.2          & 39.4          & 45.6          & \textbf{90.8} & 80.3          & 84.4          & 93.6          \\
                            & w/                                                                                        & \textbf{48.4} & \textbf{56.7} & \textbf{80.4} & \textbf{39.5} & \textbf{46.1} & 90.3          & \textbf{80.7} & \textbf{85.4} & \textbf{93.9} \\ \hline
\multirow{2}{*}{PointNet++} & w/o                                                                                       & 55.0          & 64.3          & 84.4          & 32.5          & 39.2          & \textbf{85.2} & \textbf{81.9} & \textbf{86.9} & \textbf{94.2} \\
                            & w/                                                                                        & \textbf{57.5} & \textbf{68.2} & \textbf{85.5} & \textbf{32.8} & \textbf{39.6} & 84.2          & 79.5          & 82.8          & 94.1          \\ \hline
\end{tabular}
\end{table*}

\begin{table*}[]
\centering
\caption{Per-class quantitative results on S3DIS.}
\label{table3}
\begin{tabular}{c|c|ccccccccccccc}
\hline
Backbone                    & \begin{tabular}[c]{@{}c@{}}Uncertainty-\\guided\\ learning\end{tabular} & ceiling       & floor         & wall          & beam & column        & window        & door          & table         & chair         & sofa          & bookcase      & board         & clutter       \\ \hline
\multirow{2}{*}{PointNet}   & w/o                                                                   & \textbf{91.4} & \textbf{98.1} & \textbf{73.6} & \textbf{0.1}  & 2.0           & 46.7          & 13.0          & 58.6          & 55.9          & 22.3          & 52.4          & 43.5          & 36.1          \\
                            & w/                                                                    & 89.7          & 97.5          & 72.9          & 0.0  & \textbf{10.1} & \textbf{49.4} & \textbf{18.7} & \textbf{62.4} & \textbf{61.5} & \textbf{23.7} & \textbf{54.5} & \textbf{51.8} & \textbf{37.1} \\ \hline
\multirow{2}{*}{PointNet++} & w/o                                                                   & \textbf{91.2} & \textbf{97.6} & 76.7          & 0.0  & \textbf{11.3} & 55.0          & 10.0          & \textbf{73.3} & 80.0          & 43.6          & 66.9          & \textbf{64.2} & \textbf{45.4} \\
                            & w/                                                                    & 91.1          & 97.5          & \textbf{78.1} & 0.0  & 8.7           & \textbf{58.3} & \textbf{26.4} & 71.6          & \textbf{82.4} & \textbf{57.6} & \textbf{67.5} & 60.1          & 48.1          \\ \hline
\end{tabular}
\end{table*}

\subsection{Evaluation Metrics}
We introduce the following metrics to sufficiently evaluate the performance of our method.
\subsubsection{Semantic Segmentation}
The commonly used metrics, the overall accuracy (oAcc), the mean accuracy of each class (mAcc), and the mean IoU of each class (mIoU), are used to evaluate the performance of semantic segmentation.
\subsubsection{Predictive Uncertainty Estimation}

\begin{figure}[ht]
    \centering
    \includegraphics[width=0.47\textwidth]{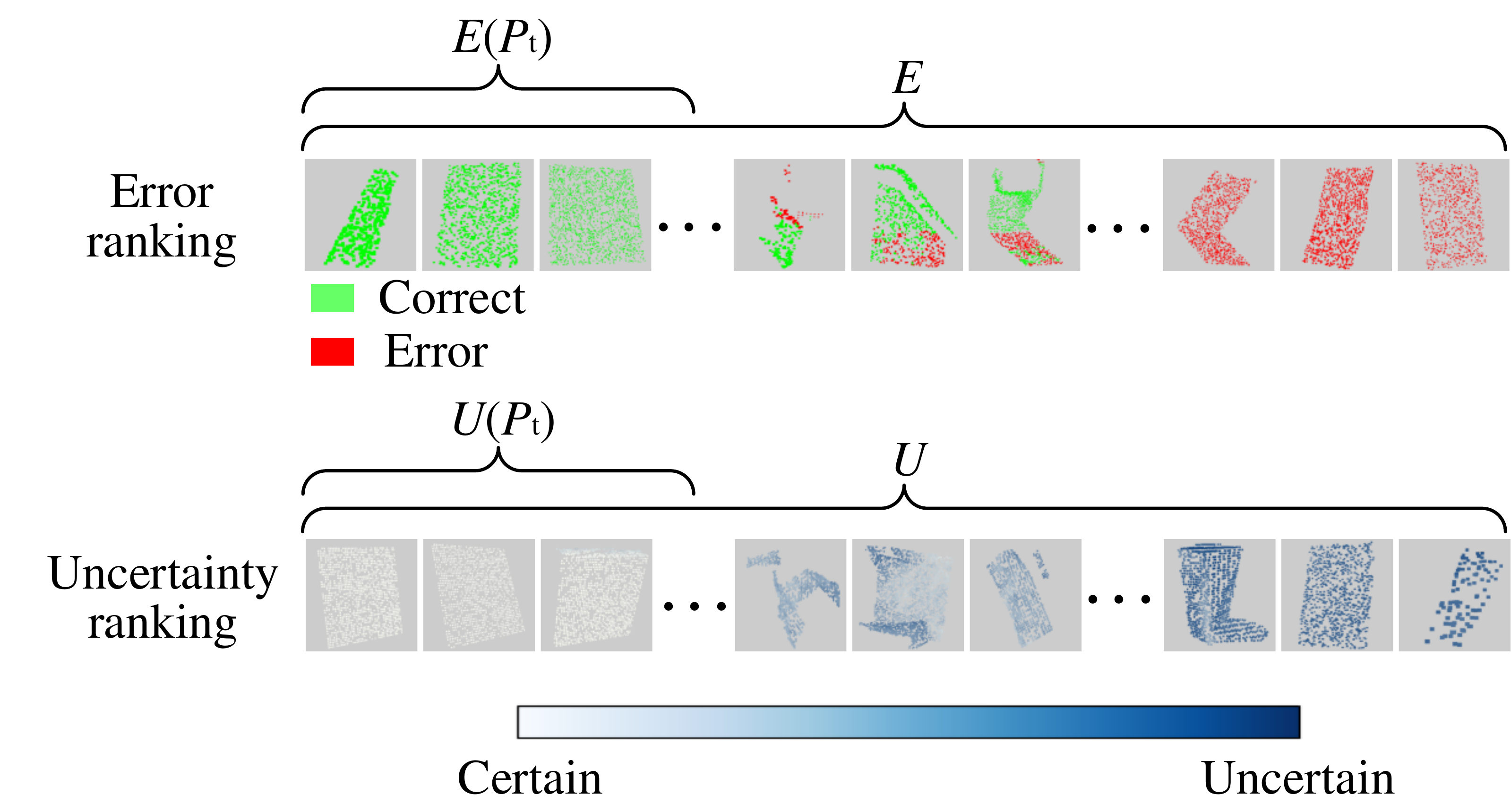}
    \caption{Explanation of Ranking IoU. The error ranking sequence is created by ranking the voxels in ascending order of prediction error, while the uncertainty ranking sequence in ascending order of mean uncertainty. Ranking IoU measures the similarity of these two sequences by calculating the intersection over the union.}
    \label{fig6}
\end{figure}

\begin{itemize}
  \item \textbf{Precision-Recall curve (PR-curve)}. A fine-grained metric inspired by \cite{Kendall2017}. This metric is used to represent the point-level relationship between predictive uncertainty and predictive precision. Points are sorted in increasing order of the uncertainty value. The vertical axis shows the prediction precision, expressed by oAcc, mAcc, or mIoU. The horizontal axis represents the percentage of recall points, which means we keep how many certain points to calculate the precision. PR-curve generated by a reliable uncertainty estimation method should decrease monotonically.
  \item \textbf{Ranking IoU}. Tasks, such as robotic grasping, focus on considering the confidence degree of prediction results within a certain spatial range. Inspired by \cite{Huang2018ECCV}, we use a coarse-grained metric called Ranking IoU to achieve voxel-level evaluation for the performance of uncertainty estimation. Occupancy voxels, transformed from the point cloud, are individually ranked according to the mean prediction error and the mean uncertainty of included points. It results in two ranking sequences: error ranking sequence \emph{E} and uncertainty ranking sequence $U$ shown in Fig. \ref{fig6}. Given a percentile threshold $P_t$, we retrieve the sequences to generate two subsequences $E({P_t}) \subset E$ and $U({P_t}) \subset U$. Ranking IoU is defined to quantify the similarity of these two subsequences, thus reflecting the correlation between predictive error and predictive uncertainty.
  \begin{equation}
  {\rm{Ranking\ IoU  =  }}\frac{{E({P_t}) \cap U({P_t})}}{{E({P_t}) \cup U({P_t})}}\label{5}
  \end{equation}
  \item \textbf{Efficiency}. We use the Runtime of inference to denote the efficiency.
\end{itemize}

\subsection{Comparison with Baselines}
\subsubsection{Baselines}
Backbone PointNet(++) without uncertainty-guided learning is chosen as the baseline in semantic segmentation comparison to verify our uncertainty-guided learning on semantic segmentation improvement. The widely used MC dropout is considered as the baseline in predictive uncertainty estimation comparison to prove the effectiveness of our method on quantifying the confidence degree of predictive results.
\subsubsection{Semantic Segmentation}

\begin{figure*}[]
    \centering
    \subfigure[]{
    \includegraphics[width=0.46\textwidth]{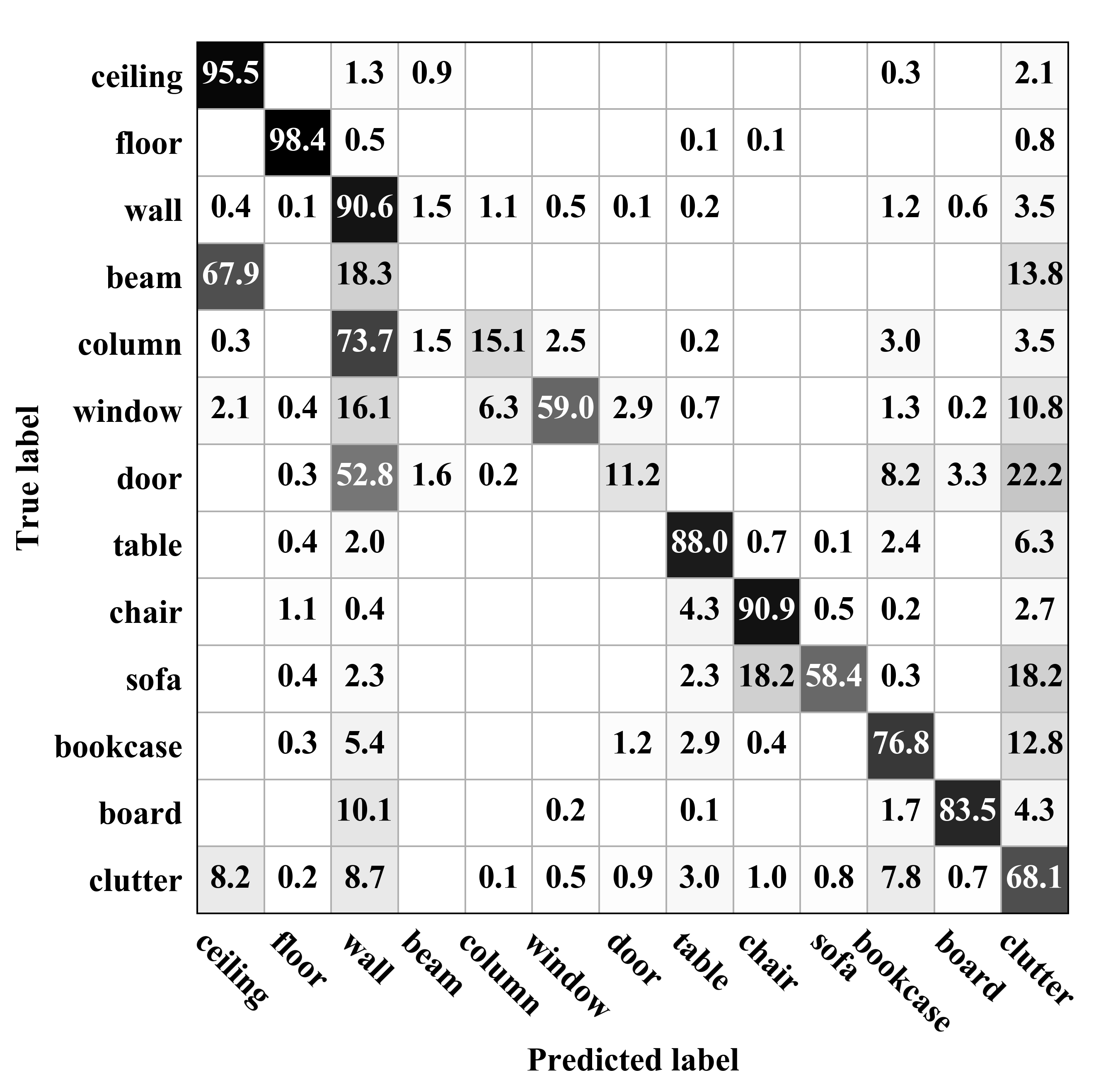}
    }
     \centering
    \subfigure[]{
    \includegraphics[width=0.46\textwidth]{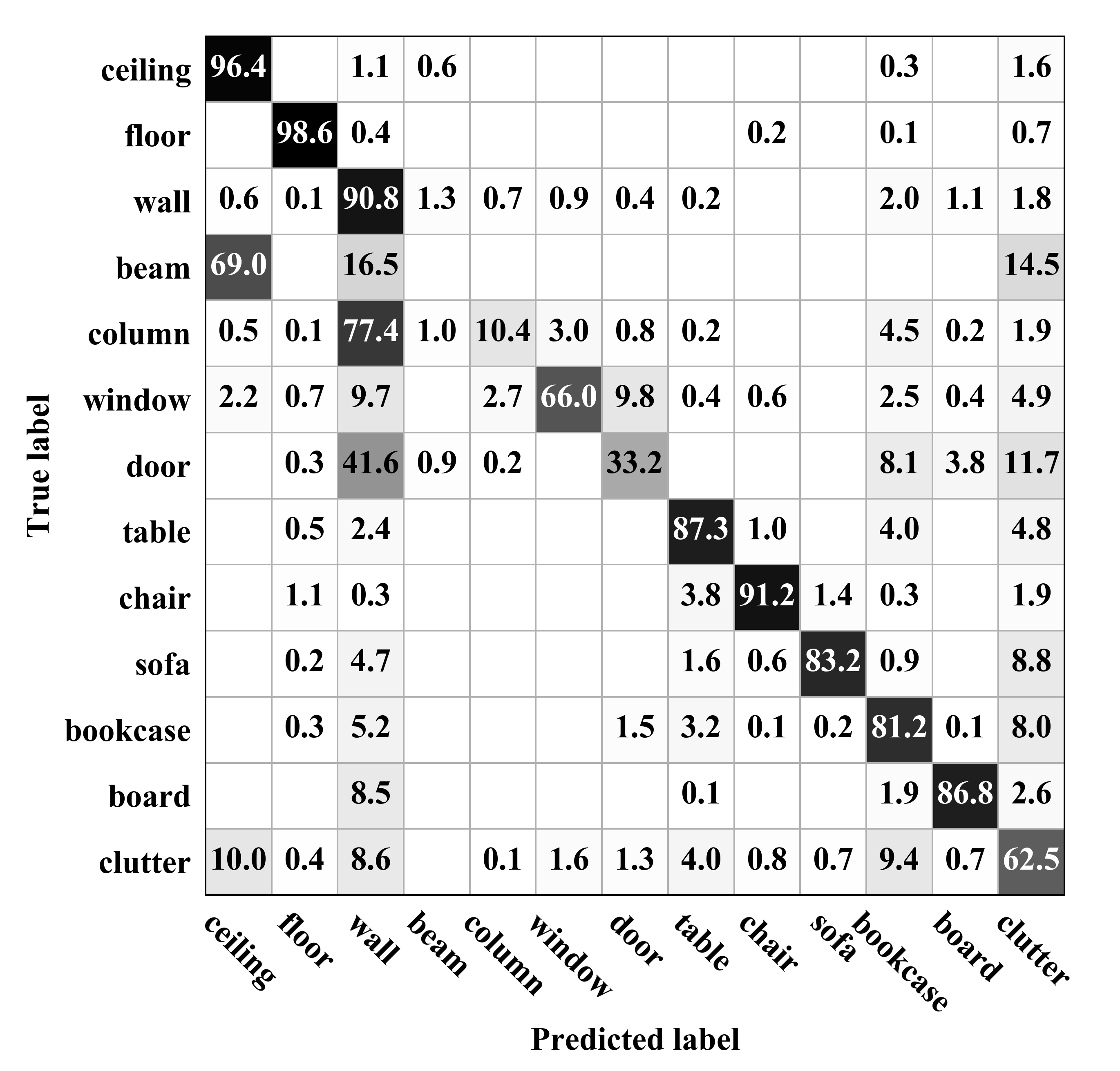}
    }
    \caption{Normalized confusion matrices. Take \emph{door} and \emph{window} as an example; many \emph{door} points are misclassified as the \emph{wall} by (a) PointNet++ w/o uncertainty-guided learning. (b) Uncertainty-guided learning effectively improves this situation.}
    \label{fig7}
\end{figure*}

\textbf{Results on S3DIS}. First, we consider the S3DIS benchmark. Table \ref{table2} shows that our method with uncertainty-guided learning outperforms the baseline method in terms of mIoU, mAcc, and oAcc (Uncertainty-guided PointNet: +2.8 mIoU, +1.7 mAcc, and +0.2 oAcc; Uncertainty-guided PointNet++: +2.5 mIoU, +3.9 mAcc, and +1.1 oAcc). The result verifies the effectiveness of our proposed uncertainty-guided learning on the real-world indoor dataset. Compare to the noticeable increase in mIoU and mAcc, the increase in oAcc is relatively small. It indicates that uncertainty-guided learning cannot dramatically improve the overall accuracy of semantic modeling for S3DIS. However, the method promotes better segmentation results on the easily-confused points, resulting in a more precise segmentation result for the point cloud. The per-class IoU and normalized confusion matrices are used to explore the details of segmentation performance improvement further.

From the per-class IoU illustrated in Table \ref{table3}, we can see that the segmentation quality of many class points has a significant improvement, particularly for points of \emph{door} and \emph{window}. In the indoor point cloud, geometric feature differences between the \emph{door}/\emph{window} and the \emph{wall} are small, thus leading to ambiguous boundaries between them and posing challenges to the manual annotation. Points with noisy labels are easily generated near these boundaries, affecting the segmentation performance. Our uncertainty-guided learning method penalizes these noisy points by integrating the quantified aleatoric uncertainty into the loss, achieving robustness to the classification of easily-confused points, as shown in normalized confusion matrices (Fig. \ref{fig7}).

\textbf{Results on NPM3D and ShapeNet}. We further evaluate the proposed uncertainty-guided learning on ShapeNet and NPM3D. Different from the obvious improvement on S3DIS as illustrated in Table \ref{table2}, our method presents a slight improvement in the segmentation of NPM3D (Uncertainty-guided PointNet: +0.1 mIoU, +0.5 mAcc, and -0.5 oAcc; Uncertainty-guided PointNet++: +0.3 mIoU, +0.4 mAcc, and -1.0 oAcc) and a not good segmentation performance on ShapeNet (Uncertainty-guided PointNet: +0.4 mIoU, +1.0 mAcc, and +0.3 oAcc; Uncertainty-guided PointNet++: -2.4 mIoU, -4.1 mAcc, and -0.1 oAcc).

ShapeNet is a synthetic dataset sampled from CAD models, which does not rely on manual annotation and has no points with noise labels. NPM3D is an urban outdoor point cloud dataset with obvious boundaries between different objects. Unlike S3DIS, the significant interclass difference of NPM3D reduces the probability of generating noisy points in the manual annotation. For these point clouds with low-noise or no-noise labels, our uncertainty-guided learning method has limited improvement.

In summary, different noise distribution of datasets leads to different degrees of segmentation performance improvement. Even though our method cannot work well in each dataset, our proposed method, which models the predictive distribution in a single forward propagation, can promote PCSS applications guided by different kinds of uncertainties.

\subsubsection{Predictive Uncertainty Estimation}

\begin{table*}[]
\centering
\caption{Evaluation results of uncertainty estimation in terms of Ranking IoU. S3DIS and NPM3D are parcellated into about two hundred occupancy voxels; every object in the ShapeNet is treated as a voxel for Ranking IoU evaluation.}
\label{table4}
\begin{tabular}{c|c|cccc|cccc|cccc}
\hline
\multirow{3}{*}{\begin{tabular}[c]{@{}c@{}}Acquisition\\ functions\end{tabular}}                                                      & \multirow{3}{*}{Method} & \multicolumn{4}{c|}{\multirow{2}{*}{S3DIS}} & \multicolumn{4}{c|}{\multirow{2}{*}{NPM3D}} & \multicolumn{4}{c}{\multirow{2}{*}{ShapeNet}} \\
                                                                               &                         & \multicolumn{4}{c|}{}                       & \multicolumn{4}{c|}{}                       & \multicolumn{4}{c}{}                          \\ \cline{3-14}
                                                                               &                         & $P_t$=10\%   & 30\%   & 50\%  & 70\%  & $P_t$=10\%   & 30\%   & 50\%  & 70\%  & $P_t$=10\%   & 30\%   & 50\%   & 70\%   \\ \hline
\multirow{2}{*}{\begin{tabular}[c]{@{}c@{}}PE\end{tabular}} & MC                & 63.2      & 76.3      & 81.8     & 86.2     & 57.8      & 71.9      & 81.8     & 88.4     & 52.4      & 73.4      & 86.0      & 86.0      \\
                                                                               & Ours                    & 63.2      & 71.2      & 80.6     & 85.4     & 58.2      & 74.6      & 85.3     & 91.2     & 52.4      & 73.4      & 86.1      & 84.8      \\ \hline
\multirow{2}{*}{Mean STD}                                                      & MC                & 57.9      & 74.6      & 77.8     & 86.2     & 48.8      & 70.3      & 79.1     & 87.5     & 52.4      & 62.5      & 73.8      & 80.7      \\
                                                                               & Ours                    & 57.9      & 69.5      & 76.5     & 85.4     & 54.7      & 67.9      & 78.3     & 88.5     & 47.6      & 64.1      & 76.9      & 80.1      \\ \hline
\end{tabular}
\end{table*}

We compare with the MC dropout baseline (MC for short) in terms of different metrics to verify the effectiveness and efficiency of our proposed uncertainty estimation method.

\begin{figure}[]
    \centering
    \subfigure[]{
    \includegraphics[width=0.3\textwidth]{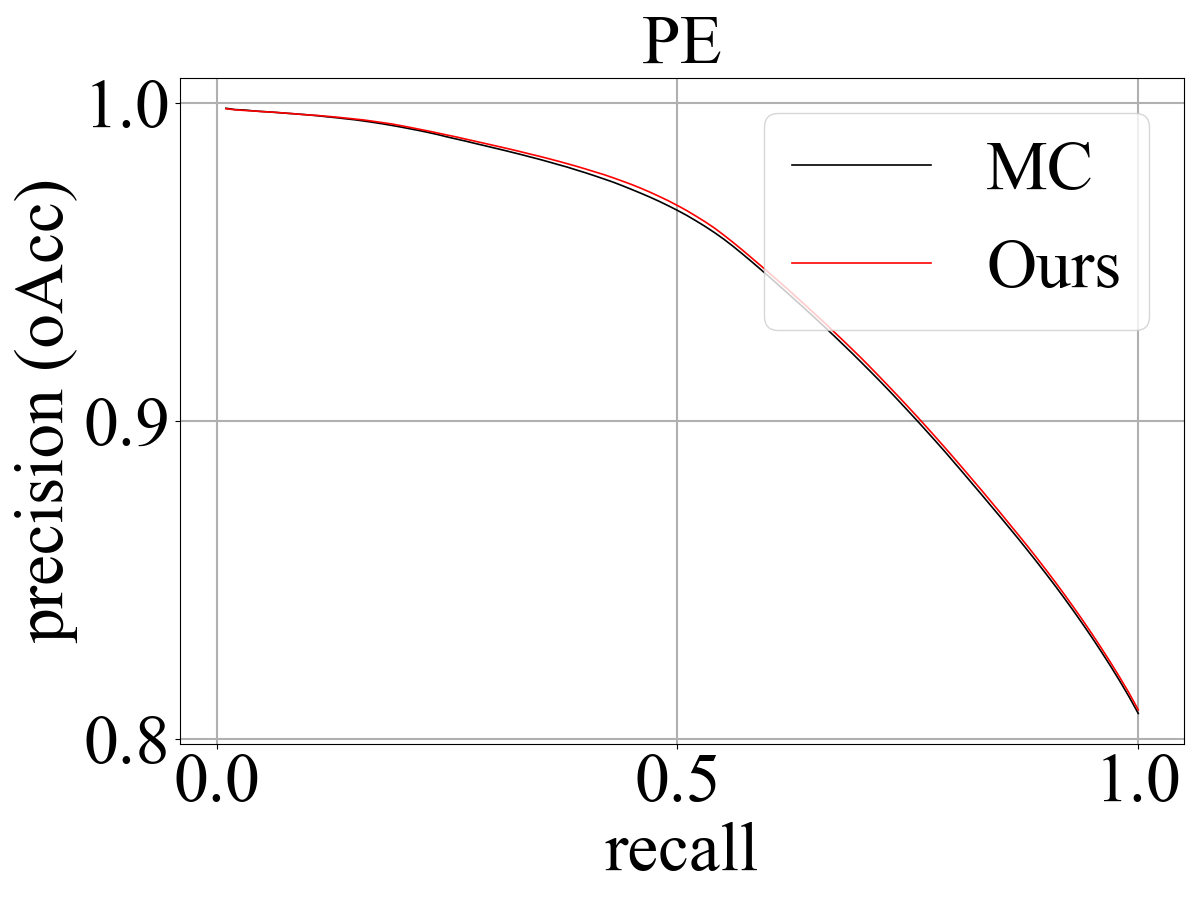}
    }
     \centering
    \subfigure[]{
    \includegraphics[width=0.3\textwidth]{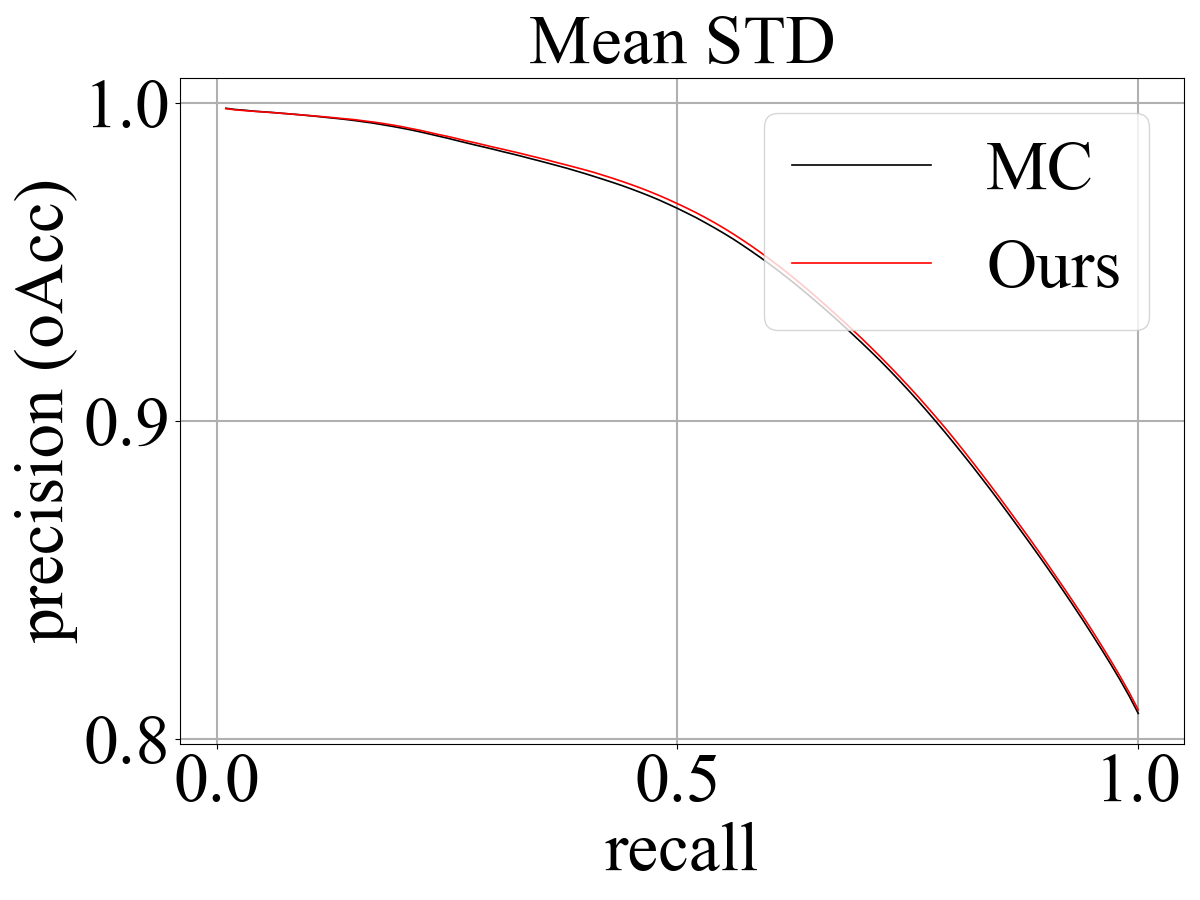}
    }
    \caption{Uncertainty estimation PR-curves of S3DIS semantic segmentation using different acquisition functions. Our method presents the same curve trend as the MC dropout, and the overall accuracy drops steadily with the increase of uncertain points. (a) PE as the acquisition function. (b) Mean STD as the acquisition function.}
    \label{fig8}
\end{figure}

\textbf{Point-level evaluation}. The PR-curve, generated by the uncertainty-aware PointNet on S3DIS, is shown in Fig. \ref{fig8} as an example. For one thing, the curves generated by our method and the MC dropout both have an apparent downward trend with the increase of uncertainty, from 100\% accuracy to the final prediction accuracy. It indicates the strong correlation between predictive accuracy and predictive uncertainty. For another thing, our method reveals a high-ranked consistency with MC dropout, verifying the effectiveness of the proposed method.

\textbf{Voxel-level evaluation}. The evaluation results in terms of Ranking IoU using PointNet as backbone are illustrated in Table \ref{table4}. Ranking IoU is generally large and rises with increasing $P_t$. Besides, the results obtained by our method are very close to those obtained by MC dropout. From the results, we can see that $E(P_t)$ is very similar to $U(P_t)$, and the similarity increases with $P_t$. Taking $P_t=10\%$ as an example, the Ranking IoU of these two subsequences is as high as about 60\%. It indicates that voxels with small uncertainties are more likely to have small prediction errors, and our method well establishes the voxel-level relationship between predictive accuracy and uncertainty.

To sum up, our proposed method based on a single forward pass has a comparable performance of uncertainty estimation with the MC dropout based on multiple forward passes in terms of different levels of metrics. Next, we analyze and verify the efficiency advantage of our method.

\textbf{Runtime analysis}. The comparison of our method and MC dropout in the runtime of uncertainty estimation is shown in Table \ref{table5}. Our method is about 5x faster than MC dropout with five-times sampling and 10x faster than MC dropout with ten-times sampling.

\begin{table}[]
\centering
\caption{Mean runtime in seconds for uncertainty estimation.}
\label{table5}
\begin{tabular}{c|c|c|cc}
\hline
\multirow{3}{*}{Backbone}                                              & \multirow{3}{*}{\begin{tabular}[c]{@{}c@{}}Sampling\\ times\end{tabular}} & \multirow{3}{*}{Method} & \multicolumn{2}{c}{Mean runtime(s)}                                                                                                                              \\
                                                                       &                                                                           &                         & \multirow{2}{*}{\begin{tabular}[c]{@{}c@{}}a room\\ in S3DIS\end{tabular}} & \multirow{2}{*}{\begin{tabular}[c]{@{}c@{}}a cropped scene\\ in NPM3D\end{tabular}} \\
                                                                       &                                                                           &                         &                                                                            &                                                                                     \\ \hline
\multirow{4}{*}{PointNet}                                              & \multirow{2}{*}{5}                                                        & MC                      & 155.0                                                                      & 308.6                                                                               \\
                                                                       &                                                                           & Ours                    & \textbf{34.8}                                                              & \textbf{71.3}                                                                       \\ \cline{2-5}
                                                                       & \multirow{2}{*}{10}                                                       & MC                      & 302.8                                                                      & 647.5                                                                               \\
                                                                       &                                                                           & Ours                    & \textbf{38.5}                                                              & \textbf{74.4}                                                                       \\ \hline
\multirow{4}{*}{\begin{tabular}[c]{@{}c@{}}PointNet\\ ++\end{tabular}} & \multirow{2}{*}{5}                                                        & MC                      & 277.5                                                                      & 677.5                                                                               \\
                                                                       &                                                                           & Ours                    & \textbf{60.4}                                                              & \textbf{148.7}                                                                      \\ \cline{2-5}
                                                                       & \multirow{2}{*}{10}                                                       & MC                      & 634.6                                                                      & 1338.5                                                                              \\
                                                                       &                                                                           & Ours                    & \textbf{62.3}                                                              & \textbf{152.5}                                                                      \\ \hline
\end{tabular}
\end{table}

It is noteworthy that MC dropout's runtime is positively correlated with the sampling time. Conversely, our method's runtime does not grow much with the increase of sampling times. Our method estimates the predictive uncertainty through neighborhood information aggregation. The increased sampling times lead to an increased number of searching neighbors, just causing a slight growth in the runtime. However, MC dropout achieves uncertainty estimation depending on repeated inferences, and the runtime scales linearly with the sampling times.

\subsection{Qualitative Results}

\begin{figure*}[]
    \centering
    \includegraphics[width=0.6\textwidth]{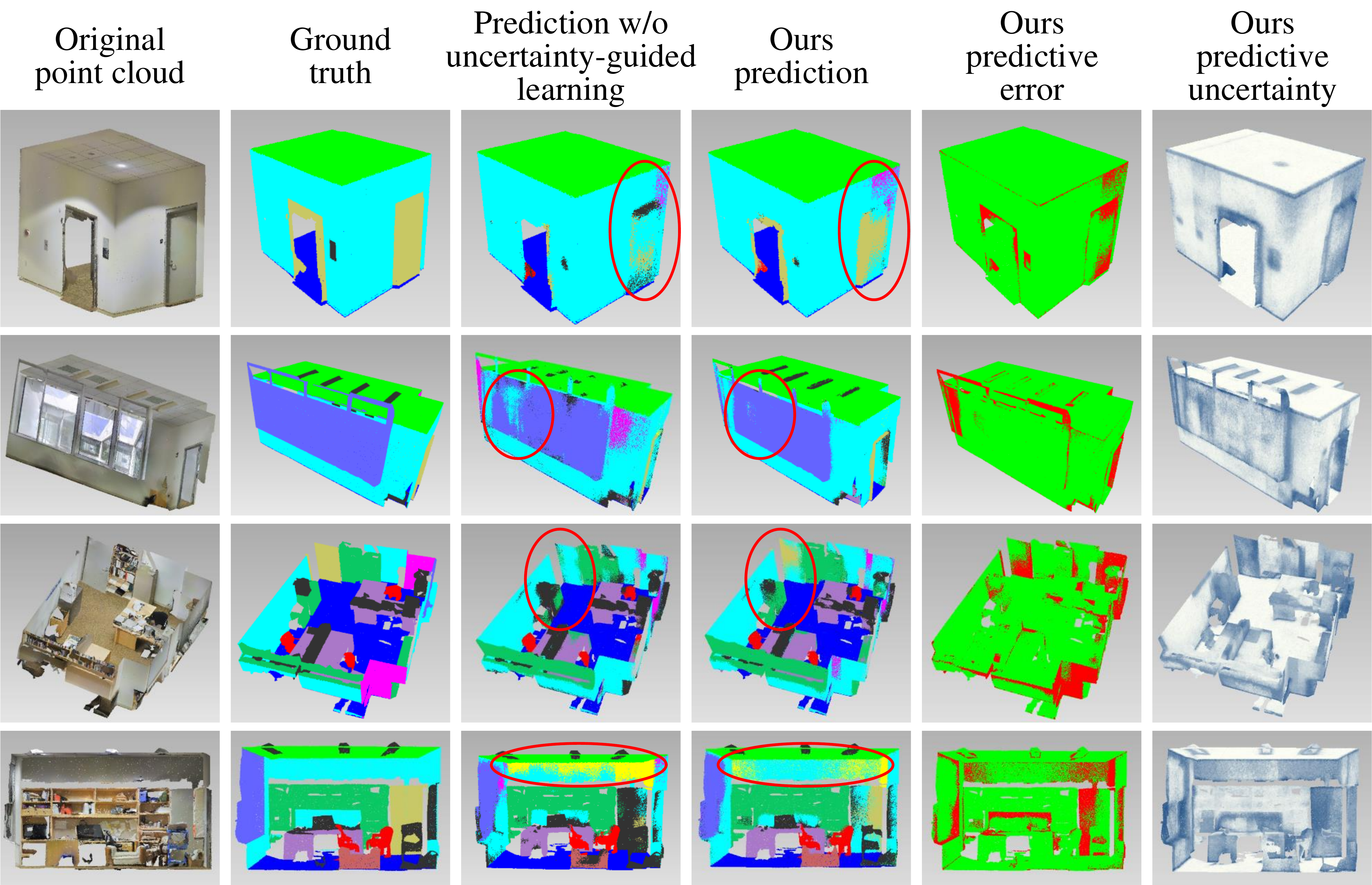}
    \caption{Visualization of uncertainty-aware PCSS on several scenes in S3DIS, with PointNet++ as the backbone. Different colors represent different semantic labels in the middle three figures, while white to dark blue indicate increasing uncertainty in the right figure. Our method improves the segmentation result, as well as outputs the predictive uncertainty.}
    \label{fig9}
\end{figure*}

Fig. \ref{fig9} illustrates the effectiveness of our uncertainty-aware PCSS on S3DIS. On the one hand, our method optimizes the prediction of easily-confused points in the red circle. With the scene on the first column as an example, lots of \emph{Door} points are misclassified as \emph{Wall} in the original prediction because of the small interclass variance. Our uncertainty-guided learning promotes \emph{Door} points to be correctly classified, and the final segmentation result has been improved significantly. On the other hand, the misclassified points usually present a high predictive uncertainty.

\begin{figure}[]
    \centering
    \includegraphics[width=0.38\textwidth]{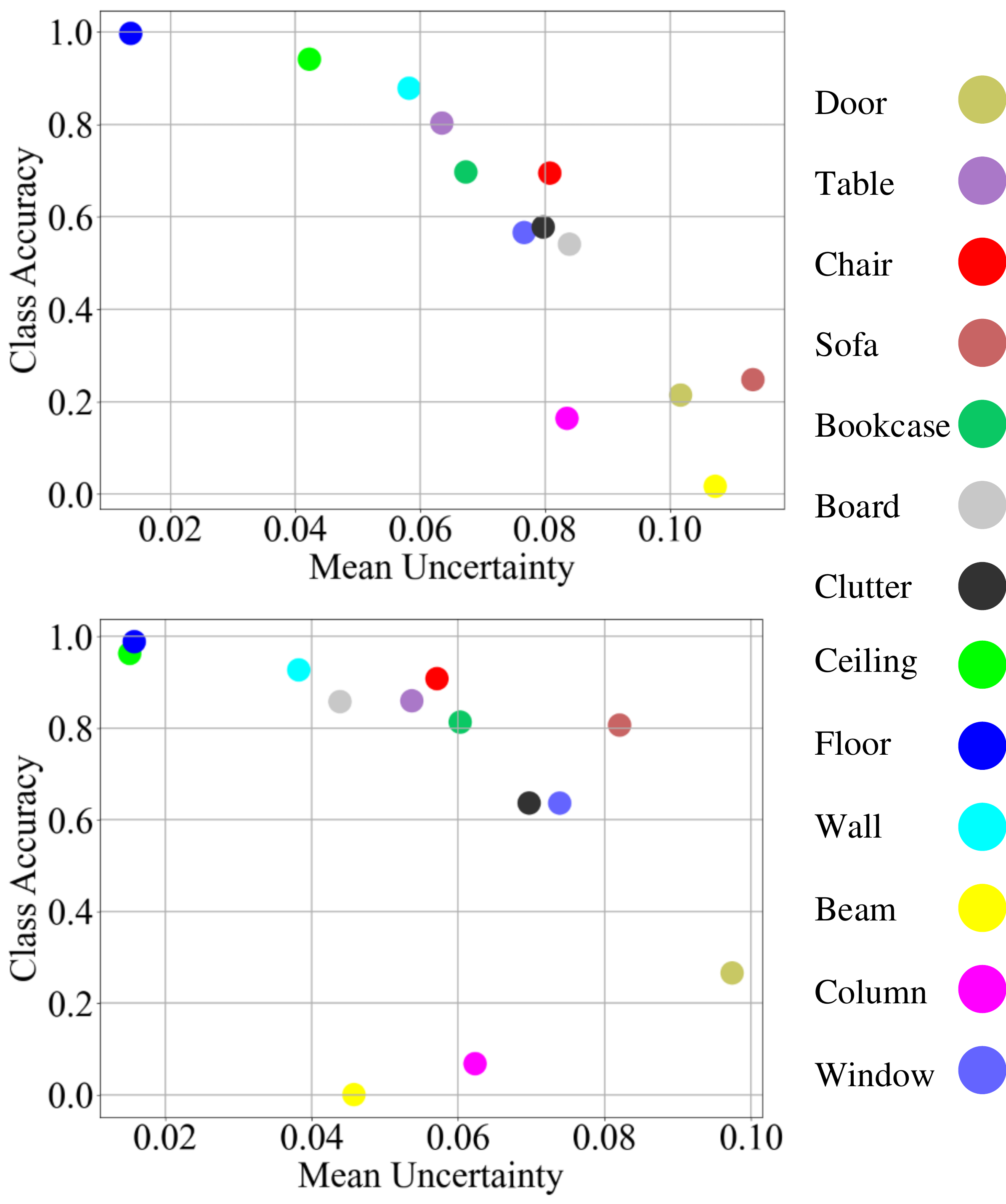}
    \caption{The relationship between class accuracy and mean uncertainty on S3DIS. The top figure represents the relationship with PointNet as the backbone, while the bottom one with PointNet++ as the backbone.}
    \label{fig10}
\end{figure}

\begin{figure*}[]
    \centering
    \includegraphics[width=0.6\textwidth]{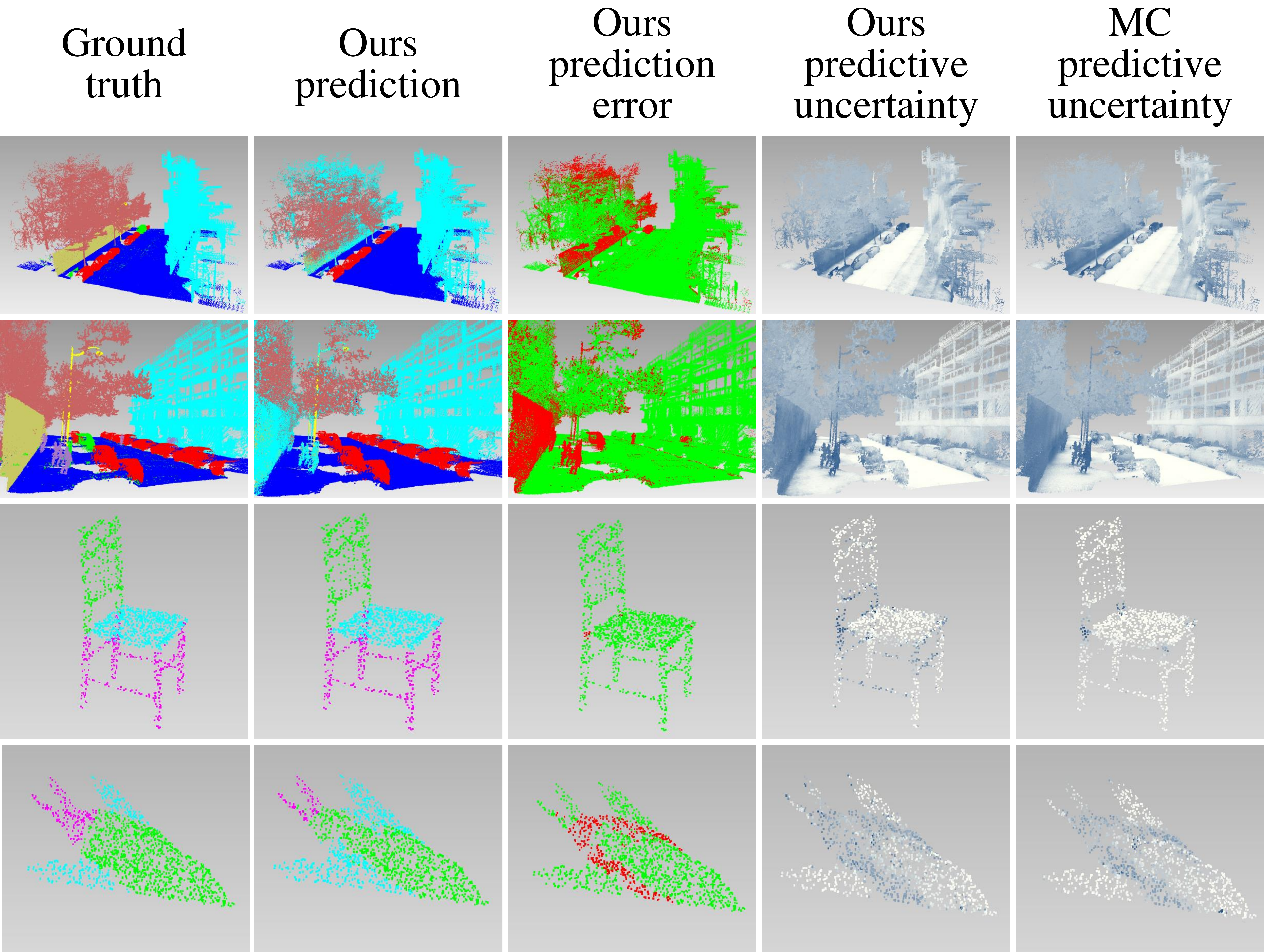}
    \caption{Uncertainty estimation on NPM3D and ShapeNet using our method and MC dropout. Misclassified points (marked in red in the middle figure) are more likely to be with large predictive uncertainty.}
    \label{fig11}
\end{figure*}

Fig. \ref{fig10} denotes the per-class relationship between predictive accuracy and predictive uncertainty on S3DIS. Besides, Fig. \ref{fig11} shows the uncertainty estimation visualization on other datasets. These figures show that our method accurately quantifies the predictive uncertainty, which has a strong inverse relationship with the predictive accuracy. Points with a large uncertainty are more likely to be misclassified. Taking \emph{Bollard} in NPM3D as an example, the backbone PointNet++ misclassifies many points, and our uncertainty-aware method shows a lack of confidence in the predictive results. The qualitative results of our method are similar to those of MC dropout, proving the effectiveness of our predictive distribution establishment method based on a single forward propagation.

\subsection{Ablation Study}

We explore the advantages of several important designs by individually removing or changing them and quantify their influence on semantic segmentation and uncertainty estimation.

\subsubsection{Effect of uncertainty-guided loss}

In uncertainty-guided learning, the designed loss function guides the model with dropout layers for better segmentation performance. However, the dropout layer itself can also improve the generalization ability of the model. To judge where the performance improvement comes from, we replace the uncertainty-guided cross-entropy (UGCE) loss function with the traditional cross-entropy (CE) loss and evaluate the influence on semantic segmentation. As illustrated in Table \ref{table6}, our uncertainty-guided learning model achieves superior performance over the model with the traditional CE loss. (Uncertainty-guided PointNet: +1.0 mIoU, +0.3 mAcc, and +0.8 oAcc; Uncertainty-guided PointNet++: 1.6 mIoU, -0.9 mAcc, and 1.6 oAcc). It suggests that although dropout layers help to improve the segmentation performance, our uncertainty-guided learning promotes further model optimization.

\begin{table}[]
\centering
\caption{Ablation study of the uncertainty-guided loss function on S3DIS. These two models are with the same configurations and dropout layers.}
\label{table6}
\begin{tabular}{c|c|ccc}
\hline
Backbone                    & \begin{tabular}[c]{@{}c@{}}Loss\\ function\end{tabular} & mIoU (\%) & mAcc (\%) & oAcc (\%) \\ \hline
\multirow{2}{*}{PointNet}   & CE                                                      & 47.4      & 56.4      & 79.6      \\
                            & UGCE                                                    & \textbf{48.4}      & \textbf{56.7}      & \textbf{80.4}      \\ \hline
\multirow{2}{*}{PointNet++} & CE                                                      & 55.9      & 67.3      & 83.9      \\
                            & UGCE                                                    & \textbf{57.5}      & \textbf{68.2}      & \textbf{85.5}      \\ \hline
\end{tabular}
\end{table}

\subsubsection{Effect of aleatoric uncertainty weighting}

\begin{table}[]
\centering
\caption{Predictive results using different weighting values. It takes a long time to train the backbone model on the complete data set. Thus we choose to train on a subset of S3DIS containing various types of scenes.}
\label{table7}
\begin{tabular}{c|c|c}
\hline
Backbone & \begin{tabular}[c]{@{}c@{}}Weighting\\ value\end{tabular} & mIoU (\%) \\ \hline
\multirow{5}{*}{PointNet} & 0 & 29.6 \\
 & 0.3 & 29.2 \\
 & 0.5 & \textbf{30.4} \\
 & 1 & 29.3 \\
 & 2 & 28.2 \\ \hline
\multirow{5}{*}{PointNet++} & 0 & 38.2 \\
 & 0.3 & 38.7 \\
 & 0.5 & \textbf{38.8} \\
 & 1 & 35.9 \\
 & 2 & 37.8 \\ \hline
\end{tabular}
\end{table}

A proper aleatoric uncertainty weighting in the designed loss function helps to obtain better predictive results. Because of the ambiguous boundary quantified by aleatoric uncertainty between the noisy points and clean points, a too-large weighting value may mistakenly penalize the modeling of clean points, while a too-small one may penalize noisy points insufficiently. Table \ref{table7} shows the predictive results on an S3DIS subset using different weighting values. Uncertainty-guided learning with a weight value of 0.5 performs the best no matter which model is used as the backbone.

\begin{table}[]
\centering
\caption{Uncertainty estimation results with different dropout configurations and sampling times (Backbone: PointNet, Dataset: S3DIS). Con\_2 inserts the dropout layer after the first and the third MLP of the decoding net, while Con\_3 inserts the dropout layer after the second MLP.}
\label{table8}
\begin{tabular}{c|c|cccc}
\hline
\multirow{2}{*}{\begin{tabular}[c]{@{}c@{}}Dropout\end{tabular}} & \multirow{2}{*}{\begin{tabular}[c]{@{}c@{}}Sampling\\ times\end{tabular}} & \multicolumn{4}{c}{Ranking IoU} \\ \cline{3-6}
 &  & $P_t$=10\% & $P_t$=30\% & $P_t$=50\% & $P_t$=70\% \\ \hline
\multirow{4}{*}{Con\_1}  & 5 & 63.2 & 72.9 & 81.6 & 84.7 \\
 & 10 & 63.2 & 71.2 & 80.6 & 85.4 \\
 & 15 & 57.9 & 72.9 & 81.6 & 86.1 \\
 & 20 & 41.4 & 74.7 & 84.9 & 87.3 \\ \hline
Con\_2 & 10 & 63.2 & 72.9 & 82.8 & 86.2 \\ \hline
Con\_3 & 10 & 57.9 & 76.3 & 80.8 & 84.8 \\ \hline
\end{tabular}
\end{table}

\subsubsection{Effect of sampling times}

Table \ref{table8} illustrates the effect of sampling times and model sampling configuration for the uncertainty estimation. Ranking IoU does not show much difference when the sampling times are set to 5, 10, and 15. It indicates that the predictive distribution of a point can be well established by aggregating the probabilistic outputs of a certain range of neighbors. However, it decreases obviously at $P_t$=10\% when the sampling time is set to 20. With the increase of neighborhood points, the difference of geometric features between points will become larger, thus leading to the inconsistency of predictive results. Our predictive distribution establishment method based on spatial aggregation cannot approximate the traditional method based on temporal aggregation, thus leading to a poor uncertainty estimation result.

\subsubsection{Effect of dropouts}

We design two variants (Con\_2, Con\_3) of our original dropout configuration (Con\_1) and explore the effect of dropout layers on uncertainty estimation. The decoder part of the backbone PointNet is mainly composed of MLPs (512, 256, 256,128, \emph{k}), where numbers in bracket are layer sizes, and \emph{k} refers to the number of classes. Con\_1 represents dropout layers after every MLP of the decoding net, while Con\_2 and Con\_3 represent dropout layers after partial MLPs. From Table \ref{table8}, we observe qualitatively that all these configurations produce similar-looking predictive uncertainty outputs, which shows that the uncertainty estimation is not sensitive to the design of dropout layers.

\section{Conclusion}

Our paper presents a framework to achieve efficient uncertainty-aware PCSS based on a single forward pass. This work focuses on efficiently leveraging uncertainty to improve the segmentation performance and quantify the confidence degree of predictive results. To tackle the problem of time-consuming predictive distribution establishment in MC dropout, we design a space-dependent method to sample the model many times by performing just one stochastic forward propagation and work with the NSA to generate the predictive distribution of each point.

Our work achieves uncertainty-aware segmentation by working with existing PCSS networks. It avoids the model overfitting by penalizing noisy points in observations and outputs the results' confidence degree. Experimental results prove the effectiveness and timeliness of our method. It will promote the application of real-world PCSS tasks.


%

\ifCLASSOPTIONcompsoc
  \section*{Acknowledgments}
\else
  \section*{Acknowledgment}
\fi

This work was supported partly by the Fundamental Research Funds for the Central Universities(Grant No. 2020XD-A04-1), partly by the National Natural Science Foundation of China (Grant No. 61673192), and partly by the BUPT Excellent Ph.D. Students Foundation (CX2021222).

\ifCLASSOPTIONcaptionsoff
  \newpage
\fi



%
\bibliographystyle{IEEETran}
\bibliography{mylib}

\begin{thebibliography}{10}
\providecommand{\url}[1]{#1}
\csname url@samestyle\endcsname
\providecommand{\newblock}{\relax}
\providecommand{\bibinfo}[2]{#2}
\providecommand{\BIBentrySTDinterwordspacing}{\spaceskip=0pt\relax}
\providecommand{\BIBentryALTinterwordstretchfactor}{4}
\providecommand{\BIBentryALTinterwordspacing}{\spaceskip=\fontdimen2\font plus
\BIBentryALTinterwordstretchfactor\fontdimen3\font minus
  \fontdimen4\font\relax}
\providecommand{\BIBforeignlanguage}[2]{{%
\expandafter\ifx\csname l@#1\endcsname\relax
\typeout{** WARNING: IEEEtran.bst: No hyphenation pattern has been}%
\typeout{** loaded for the language `#1'. Using the pattern for}%
\typeout{** the default language instead.}%
\else
\language=\csname l@#1\endcsname
\fi
#2}}
\providecommand{\BIBdecl}{\relax}
\BIBdecl

\bibitem{Denker1991transforming}
J.~Denker and Y.~Lecun, ``Transforming neural-net output levels to probability
  distributions,'' in \emph{Advances in Neural Information Processing Systems
  (NIPS)}, 1990, pp. 853--859.

\bibitem{MacKay1992A}
D.~J. MacKay, ``A practical bayesian framework for backpropagation networks,''
  \emph{Neural Computation}, vol.~4, no.~3, pp. 448--472, 1992.

\bibitem{Neal1995}
R.~Neal, ``Bayesian learning for neural networks.'' Ph.D. dissertation,
  University of Toronto, 1995.

\bibitem{Gal2016}
Y.~Gal and Z.~Ghahramani, ``Dropout as a bayesian approximation: Representing
  model uncertainty in deep learning,'' in \emph{International Conference on
  Machine Learning (ICML)}, 2016, pp. 1050--1059.

\bibitem{BlumHP15}
A.~Blum, N.~Haghtalab, and A.~D. Procaccia, ``Variational dropout and the local
  reparameterization trick,'' in \emph{Advances in Neural Information
  Processing Systems (NIPS)}, 2015, pp. 2575--2583.

\bibitem{Kendall2017}
A.~Kendall and Y.~Gal, ``What uncertainties do we need in bayesian deep
  learning for computer vision?'' in \emph{Advances in neural information
  processing systems}, 2017, pp. 5574--5584.

\bibitem{feng2019leveraging}
D.~{Feng}, L.~{Rosenbaum}, F.~{Timm}, and K.~{Dietmayer}, ``Leveraging
  heteroscedastic aleatoric uncertainties for robust real-time lidar 3d object
  detection,'' in \emph{IEEE Intelligent Vehicles Symposium (IV)}, 2019, pp.
  1280--1287.

\bibitem{yi2019probabilistic}
K.~{Yi} and J.~{Wu}, ``Probabilistic end-to-end noise correction for learning
  with noisy labels,'' in \emph{IEEE Conference on Computer Vision and Pattern
  Recognition (CVPR)}, 2019, pp. 7017--7025.

\bibitem{letham2019constrained}
B.~{Letham}, B.~{Karrer}, G.~{Ottoni}, and E.~{Bakshy}, ``Constrained bayesian
  optimization with noisy experiments,'' \emph{Bayesian Analysis}, vol.~14,
  no.~2, pp. 495--519, 2019.

\bibitem{yu2019uncertainty}
L.~{Yu}, S.~{Wang}, X.~{Li}, C.-W. {Fu}, and P.-A. {Heng}, ``Uncertainty-aware
  self-ensembling model for semi-supervised 3d left atrium segmentation,'' in
  \emph{Lecture Notes in Computer Science (including subseries Lecture Notes in
  Artificial Intelligence and Lecture Notes in Bioinformatics)}, 2019, pp.
  605--613.

\bibitem{ebrahimi2020uncertainty}
S.~{Ebrahimi}, M.~{Elhoseiny}, T.~{Darrell}, and M.~{Rohrbach},
  ``Uncertainty-guided continual learning with bayesian neural networks,'' in
  \emph{International Conference on Learning Representations (ICLR)}, 2020.

\bibitem{xia2020uncertainty}
Y.~{Xia}, D.~{Yang}, Z.~{Yu}, F.~{Liu}, J.~{Cai}, L.~{Yu}, Z.~{Zhu}, D.~{Xu},
  A.~L. {Yuille}, and H.~{Roth}, ``Uncertainty-aware multi-view co-training for
  semi-supervised medical image segmentation and domain adaptation,''
  \emph{Medical Image Analysis}, vol.~65, p. 101766, 2020.

\bibitem{wang2020double}
Y.~{Wang}, Y.~{Zhang}, J.~{Tian}, C.~{Zhong}, Z.~{Shi}, Y.~{Zhang}, and
  Z.~{He}, ``Double-uncertainty weighted method for semi-supervised learning.''
  in \emph{International Conference on Medical Image Computing and
  Computer-Assisted Intervention (MICCAI)}, 2020, pp. 542--551.

\bibitem{LiuLCHH20}
Z.~Liu, S.~Li, S.~Chen, Y.~Hu, and S.~Huang, ``Uncertainty aware graph gaussian
  process for semi-supervised learning,'' in \emph{AAAI Conference on
  Artificial Intelligence (AAAI)}, 2020, pp. 4957--4964.

\bibitem{guo2020deep}
Y.~{Guo}, H.~{Wang}, Q.~{Hu}, H.~{Liu}, L.~{Liu}, and M.~{Bennamoun}, ``Deep
  learning for 3d point clouds: A survey.'' \emph{IEEE Transactions on Pattern
  Analysis and Machine Intelligence (Early Access)}, 2020.

\bibitem{charles2017pointnet}
R.~Q. {Charles}, H.~{Su}, M.~{Kaichun}, and L.~J. {Guibas}, ``Pointnet: Deep
  learning on point sets for 3d classification and segmentation,'' in
  \emph{IEEE Conference on Computer Vision and Pattern Recognition (CVPR)},
  2017, pp. 77--85.

\bibitem{qi2017pointnet}
C.~R. {Qi}, L.~{Yi}, H.~{Su}, and L.~J. {Guibas}, ``Pointnet++: Deep
  hierarchical feature learning on point sets in a metric space,'' in
  \emph{Advances in Neural Information Processing Systems (NIPS)}, 2017, pp.
  5099--5108.

\bibitem{hu2020randla}
Q.~{Hu}, B.~{Yang}, L.~{Xie}, S.~{Rosa}, Y.~{Guo}, Z.~{Wang}, N.~{Trigoni}, and
  A.~{Markham}, ``Randla-net: Efficient semantic segmentation of large-scale
  point clouds,'' in \emph{IEEE Conference on Computer Vision and Pattern
  Recognition (CVPR)}, 2020, pp. 11\,108--11\,117.

\bibitem{lei2020spherical}
H.~{Lei}, N.~{Akhtar}, and A.~{Mian}, ``Spherical kernel for efficient graph
  convolution on 3d point clouds.'' \emph{IEEE Transactions on Pattern Analysis
  and Machine Intelligence (Early Access)}, 2020.

\bibitem{guo2021pct}
M.~Guo, J.~Cai, Z.~Liu, T.~Mu, R.~R. Martin, and S.~Hu, ``Pct: Point cloud
  transformer,'' \emph{Computational Visual Media}, vol.~7, no.~2, pp.
  187--199, 2021.

\bibitem{Chao2021}
C.~{Qi}, J.~{Yin}, H.~{Liu}, and J.~{Liu}, ``Neighborhood spatial aggregation
  based efficient uncertainty estimation for point cloud semantic
  segmentation,'' in \emph{International Conference on Robotics and Automation
  (ICRA)}, 2021.

\bibitem{Charles2015}
B.~Charles, C.~Julien, K.~Koray, and W.~Daan, ``Weight uncertainty in neural
  networks,'' in \emph{Proceedings of the 32nd International Conference on
  International Conference on Machine Learning (ICML)}, 2015, pp. 1613¨C--1622.

\bibitem{Hinton1993}
G.~Hinton and D.~Camp, ``Keeping the neural networks simple by minimizing the
  description length of the weights,'' in \emph{Proceedings of the sixth annual
  conference on Computational learning theory (COLT)}, 1993, pp. 5--13.

\bibitem{Barber1998}
D.~Barber and C.~M. Bishop, ``Ensemble learning in bayesian neural networks,''
  \emph{Nato ASI Series F Computer and Systems Sciences}, vol. 168, pp.
  215--238, 1998.

\bibitem{Graves2011}
A.~Graves, ``Practical variational inference for neural networks,'' in
  \emph{Advances in neural information processing systems (NIPS)}, 2011, pp.
  2348--2356.

\bibitem{zhang2021uncertainty}
J.~Zhang, D.~Fan, Y.~Dai, S.~Anwar, F.~S. Saleh, S.~Aliakbarian, and N.~Barnes,
  ``Uncertainty inspired rgb-d saliency detection,'' \emph{IEEE Transactions on
  Pattern Analysis and Machine Intelligence (Early Access)}, 2021.

\bibitem{soleimani2018scalable}
H.~{Soleimani}, J.~{Hensman}, and S.~{Saria}, ``Scalable joint models for
  reliable uncertainty-aware event prediction,'' \emph{IEEE Transactions on
  Pattern Analysis and Machine Intelligence}, vol.~40, no.~8, pp. 1948--1963,
  2018.

\bibitem{Postels2019ICCV}
J.~Postels, F.~Ferroni, H.~Coskun, N.~Navab, and F.~Tombari, ``Sampling-free
  epistemic uncertainty estimation using approximated variance propagation,''
  in \emph{Proceedings of the IEEE International Conference on Computer Vision
  (ICCV)}, October 2019, pp. 2931--2940.

\bibitem{Huang2018ECCV}
P.-Y. Huang, W.-T. Hsu, C.-Y. Chiu, T.-F. Wu, and M.~Sun, ``Efficient
  uncertainty estimation for semantic segmentation in videos,'' in
  \emph{Proceedings of the European Conference on Computer Vision (ECCV)},
  September 2018, pp. 520--535.

\bibitem{gurevich2019pairing}
P.~{Gurevich} and H.~{Stuke}, ``Pairing an arbitrary regressor with an
  artificial neural network estimating aleatoric uncertainty,''
  \emph{Neurocomputing}, vol. 350, pp. 291--306, 2019.

\bibitem{Kendall2015}
A.~Kendall, V.~Badrinarayanan, and R.~Cipolla, ``Bayesian segnet: Model
  uncertainty in deep convolutional encoder-decoder architectures for scene
  understanding,'' in \emph{British Machine Vision Conference (BMVC)}, 2017.

\bibitem{Shannon2001}
C.~E. Shannon, ``A mathematical theory of communication,'' \emph{ACM SIGMOBILE
  mobile computing and communications}, vol.~5, no.~1, pp. 3--55, 2001.

\bibitem{Gal2016phd}
Y.~Gal, ``Uncertainty in deep learning,'' Ph.D. dissertation, University of
  Cambridge, 2016.

\bibitem{depeweg2018decomposition}
S.~{Depeweg}, J.~M. {Hern¨¢ndez-Lobato}, F.~{Doshi-Velez}, and S.~{Udluft},
  ``Decomposition of uncertainty in bayesian deep learning for efficient and
  risk-sensitive learning,'' in \emph{International Conference on Machine
  Learning (ICML)}, 2018, pp. 1184--1193.

\bibitem{kampffmeyer2016semantic}
M.~{Kampffmeyer}, A.-B. {Salberg}, and R.~{Jenssen}, ``Semantic segmentation of
  small objects and modeling of uncertainty in urban remote sensing images
  using deep convolutional neural networks,'' in \emph{IEEE Conference on
  Computer Vision and Pattern Recognition Workshops (CVPRW)}, 2016, pp.
  680--688.

\bibitem{kwon2020uncertainty}
Y.~{Kwon}, J.~H. {Won}, B.~J. {Kim}, and M.~C. {Paik}, ``Uncertainty
  quantification using bayesian neural networks in classification: Application
  to biomedical image segmentation,'' \emph{Computational Statistics \& Data
  Analysis}, vol. 142, p. 106816, 2020.

\bibitem{KampffmeyerSJ16}
M.~Kampffmeyer, A.~Salberg, and R.~Jenssen, ``Semantic segmentation of small
  objects and modeling of uncertainty in urban remote sensing images using deep
  convolutional neural networks,'' in \emph{IEEE Conference on Computer Vision
  and Pattern Recognition Workshops (CVPRW)}, 2016, pp. 680--688.

\bibitem{bentaieb2017uncertainty}
A.~BenTaieb and G.~Hamarneh, ``Uncertainty driven multi-loss fully
  convolutional networks for histopathology,'' \emph{Intravascular Imaging and
  Computer Assisted Stenting, and Large-Scale Annotation of Biomedical Data and
  Expert Label Synthesis - 6th Joint International Workshops (CVII-STENT)}, pp.
  155--163, 2017.

\bibitem{Iro2016}
I.~Armeni, O.~Sener, A.~Zamir, H.~Jiang, I.~Brilakis, M.~Fischer, and
  S.~Savarese, ``3d semantic parsing of large-scale indoor spaces,'' in
  \emph{Proceedings of the IEEE Conference on Computer Vision and Pattern
  Recognition (CVPR)}, 2016, pp. 1534--1543.

\bibitem{roynard2018paris}
X.~{Roynard}, J.-E. {Deschaud}, and F.~{Goulette}, ``Paris-lille-3d: A large
  and high-quality ground-truth urban point cloud dataset for automatic
  segmentation and classification:,'' \emph{The International Journal of
  Robotics Research}, vol.~37, no.~6, pp. 545--557, 2018.

\bibitem{yi2016a}
L.~{Yi}, V.~G. {Kim}, D.~{Ceylan}, I.-C. {Shen}, M.~{Yan}, H.~{Su}, C.~{Lu},
  Q.~{Huang}, A.~{Sheffer}, and L.~{Guibas}, ``A scalable active framework for
  region annotation in 3d shape collections,'' \emph{international conference
  on computer graphics and interactive techniques}, vol.~35, no.~6, p. 210,
  2016.

\end{thebibliography}

%

\begin{IEEEbiography}[{\includegraphics[width=1in,height=1.25in,clip,keepaspectratio]{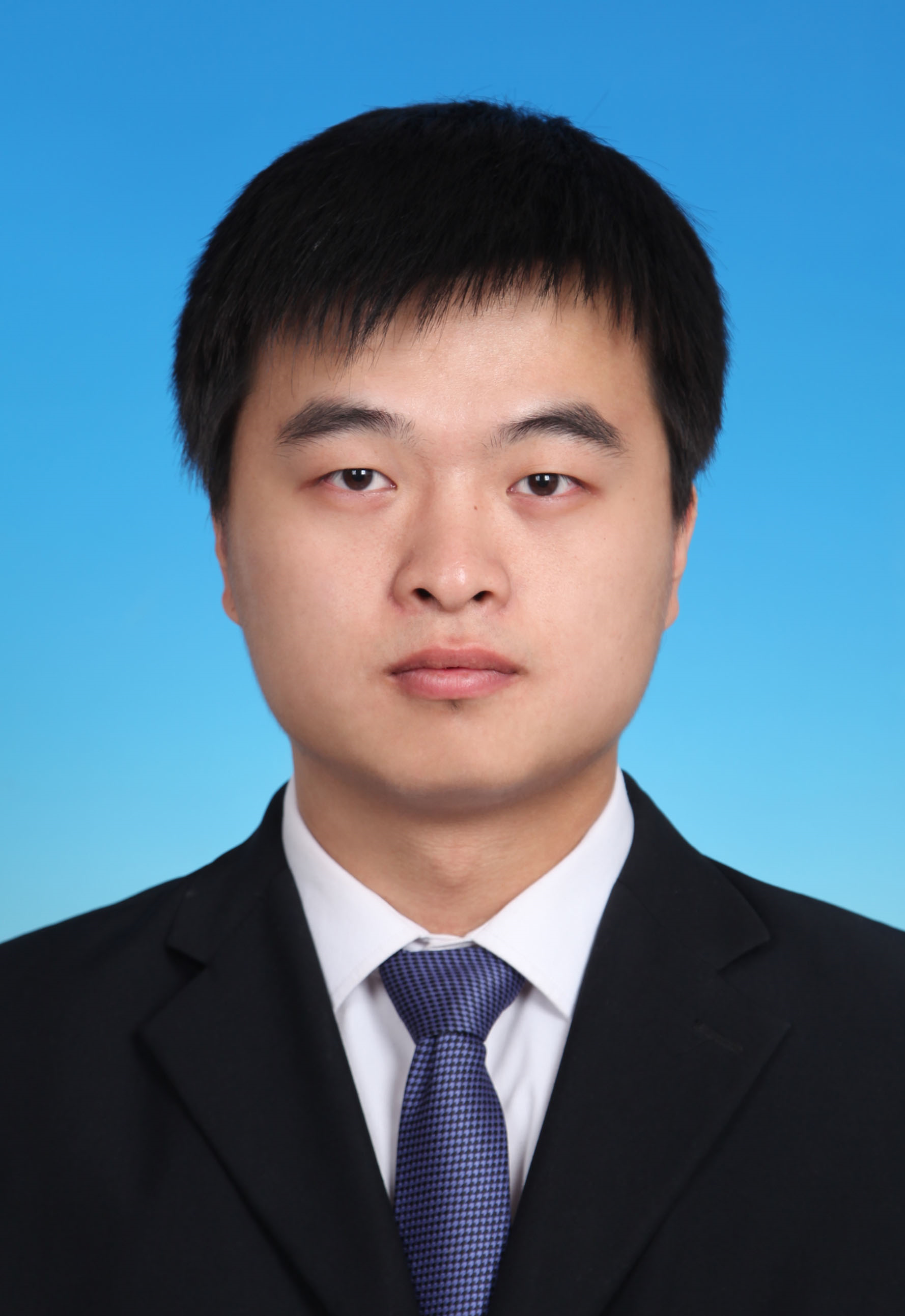}}]{Chao Qi}
received the master's degree from Beijing Jiaotong University, Beijing, China, in 2014. He is currently pursuing the Ph.D. degree with the School of Artificial Intelligence, Beijing University
of Posts and Telecommunications, Beijing, China. He is also an Assistant Research Fellow with Standard and Metrology Research Institute, China Academy of Railway Sciences Corporation Limited, Beijing, China. His research interests include machine learning and point cloud processing.
\end{IEEEbiography}

\begin{IEEEbiography}[{\includegraphics[width=1in,height=1.25in,clip,keepaspectratio]{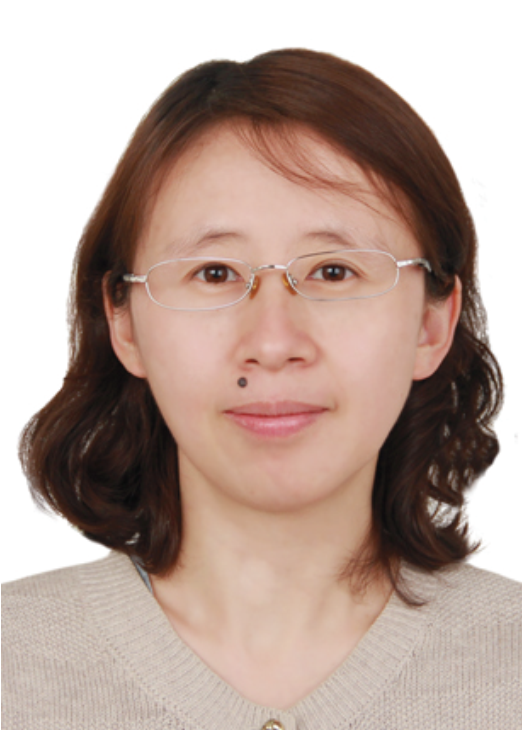}}]{Jianqin Yin}
received the Ph.D. degree from Shandong University, Jinan, China, in 2013. She currently is a Professor with the School of Artificial Intelligence, Beijing University of Posts and Telecommunications, Beijing, China. Her research interests include service robot, pattern recognition, machine learning and image processing.
\end{IEEEbiography}




\end{document}